%% file: main.tex
\documentclass[10pt,journal, compsoc]{IEEEtran}

\IEEEoverridecommandlockouts
\usepackage{cite}
\usepackage{amsmath,amssymb,amsfonts}
\usepackage{algorithmic}
\usepackage{graphicx}
\usepackage{textcomp}
\usepackage{xcolor}
\def\BibTeX{{\rm B\kern-.05em{\sc i\kern-.025em b}\kern-.08em
    T\kern-.1667em\lower.7ex\hbox{E}\kern-.125emX}}
    
\usepackage{hyperref}
\usepackage{booktabs}
\usepackage{adjustbox}
\usepackage{colortbl}
\usepackage{float}
\usepackage[most]{tcolorbox}
\usepackage{multirow,tabularx}
\newcommand{\nd}{\vspace{1mm}\noindent}
\usepackage{amsmath}
\usepackage{balance}
\usepackage{paralist}
\usepackage{amsfonts}
\usepackage{arydshln}
\usepackage[utf8]{inputenc}
\usepackage[normalem]{ulem}
\usepackage[T1]{fontenc}
\usepackage{tablefootnote}
\usepackage{soul}

\usepackage{makecell,multirow,tabu}
\usepackage{array}

\usepackage[utf8]{inputenc}
\usepackage{pst-node}
\usepackage{mathtools} 
\usepackage{enumitem}
\newlist{steps}{enumerate}{1}
\setlist[steps, 1]{wide=0pt, leftmargin=\parindent, label=Step \arabic*:, font=\bfseries}

\usepackage[referable]{threeparttablex}
\renewlist{tablenotes}{enumerate}{1}
\makeatletter
\setlist[tablenotes]{label=\tnote{\alph*},ref=\alph*,itemsep=\z@,topsep=\z@skip,partopsep=\z@skip,parsep=\z@,itemindent=\z@,labelindent=\tabcolsep,labelsep=.2em,leftmargin=*,align=left,before={\footnotesize}}
\makeatother

\newenvironment{boxedSimple}
{   
    \renewcommand{\arraystretch}{1.5}
    \begin{center}
    \vspace{+1mm}
    \begin{tabular}{|p{0.45\textwidth}|}
    \hline
        }
        { 
    \\
    \hline
    \end{tabular}
    \vspace{-6mm}
    \end{center}
}

\usepackage{listings}
\usepackage{xcolor}

\usepackage{caption}
\usepackage{tcolorbox}

\newcommand{\lcolorbox}[2]{%
  \hspace*{-\fboxsep}\colorbox{#1}{#2}%
}
\usepackage{soul}
\definecolor{mygray}{gray}{0.95}
\definecolor{mybrown}{rgb}{0.9,0.8,0.8}
\usepackage[most]{tcolorbox}

\definecolor{codegreen}{rgb}{0,0.6,0}
\definecolor{codegray}{rgb}{0.5,0.5,0.5}
\definecolor{codepurple}{rgb}{0.58,0,0.82}
\definecolor{backcolour}{rgb}{0.95,0.95,0.92}

\lstdefinestyle{mystyle}{
    backgroundcolor=\color{backcolour},   
    commentstyle=\color{codegreen},
    keywordstyle=\color{magenta},
    numberstyle=\tiny\color{codegray},
    stringstyle=\color{codepurple},
    basicstyle=\ttfamily\footnotesize\tiny,
    breakatwhitespace=false,         
    breaklines=true,                 
    captionpos=b,                    
    keepspaces=true,  
    numbers=left,                    
    showspaces=false,                
    showstringspaces=false,
    showtabs=false,                  
    tabsize=2,
    frame=lines,
    tabsize=1, 
    frame=single, 
    framerule=2pt, 
    frameround=tttt,       
    breakatwhitespace=false,          
    breaklines=true,                 
    deletekeywords={...},            
    escapeinside={\%*}{*)},           
    keepspaces=true,                 
    morekeywords={*,...},      
    numberstyle=\tiny\color{gray}, 
    rulecolor=\color{black},         
    showspaces=false,               
    showstringspaces=false,          
    showtabs=false,                  
    stepnumber=1,                    
    title=\lstname,
    xleftmargin=0.4cm, 
    xrightmargin=0.4cm, 
    escapechar=|, 
}

\definecolor{mycolor}{rgb}{0.8,0.8,0.8}
\definecolor{light-gray}{gray}{0.96}

\ifCLASSINFOpdf
\else
\fi

\begin{document}

\def\boxit#1{%
  \smash{\fboxsep=0pt\llap{\rlap{\fbox{\strut\makebox[#1]{}}}~}}\ignorespaces
}

\newcommand{\heng}[1]{\textcolor{brown}{{\it [Heng says: #1]}}}

\newcommand{\xam}[1]{\textcolor{blue}{{\it [Max says: #1]}}}

\newcommand{\amin}[1]{\textcolor{red}{{\it [Amin says: #1]}}}


\newcommand\mysubsubsection{\@startsection{subsubsection}{3}{\z@}%
                {-3.25ex\@plus -1ex \@minus -.2ex}%
                {1.5ex \@plus .2ex}%
                {\normalfont\normalsize\bfseries}}
\newcommand\myparagraph{\@startsection{paragraph}{4}{\z@}%
                {3.25ex \@plus1ex \@minus.2ex}%
                {-1em}%
                {\normalfont\normalsize\bfseries}}

\title{What Causes Exceptions in Machine Learning Applications? Mining Machine Learning-Related Stack Traces on Stack Overflow \\}
\author{Amin~Ghadesi,
        and~Maxime~Lamothe,~\IEEEmembership{Member,~IEEE,}
        and~Heng~Li
\IEEEcompsocitemizethanks{

\IEEEcompsocthanksitem Amin Ghadesi, Maxime Lamothe and Heng Li are with the Department of Computer Engineering and Software Engineering, Polytechnique Montreal, Canada.\protect\\
E-mail: \{amin.ghadesi, maxime.lamothe, heng.li\}@polymtl.ca

}
}

\markboth{Journal of TSE, April~2023}%
{Shell \MakeLowercase{\textit{et al.}}: Bare Advanced Demo of IEEEtran.cls for IEEE Computer Society Journals}
%



\IEEEtitleabstractindextext{%
\input{Text/abstract}


}

\maketitle

\IEEEdisplaynontitleabstractindextext

%
\IEEEpeerreviewmaketitle

\input{Text/introduction.tex}
\input{Text/case_study_setup}
\input{Text/results}
\input{Text/discussion}
\input{Text/threads_to_validity}

\input{Text/related_works}

\input{Text/conclusions}

\ifCLASSOPTIONcaptionsoff
  \newpage
\fi



%

\bibliographystyle{IEEEtran}
\bibliography{reference.bib}




%

\begin{IEEEbiography}                   [{\includegraphics[width=1in,height=1.25in,clip,keepaspectratio]{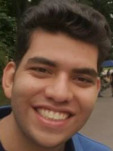}}]{Amin Ghadesi} 
Amin Ghadesi received the B.Sc. and M.Sc. degrees from the University of Mazandaran, and the Sahand University of Technology, Iran, in 2016 and 2018. Currently, he is a master's student at Polytechnique Montréal, Canada. His main research interests are data mining, machine learning, deep learning, and natural language processing tasks.    
\end{IEEEbiography}
\begin{IEEEbiography}
[{\includegraphics[width=1in,height=1.25in,clip,keepaspectratio]{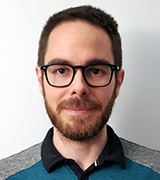}}]{Maxime Lamothe} 
Maxime Lamothe is an assistant professor at Polytechnique Montreal. He obtained his PhD from Concordia University, Montreal, Canada. Maxime uses empirical methods to study topics such as  software evolution, software APIs, software dependencies, software quality, and machine learning in software engineering.
\end{IEEEbiography}
\begin{IEEEbiography}
[{\includegraphics[width=1in,height=1.25in,clip,keepaspectratio]{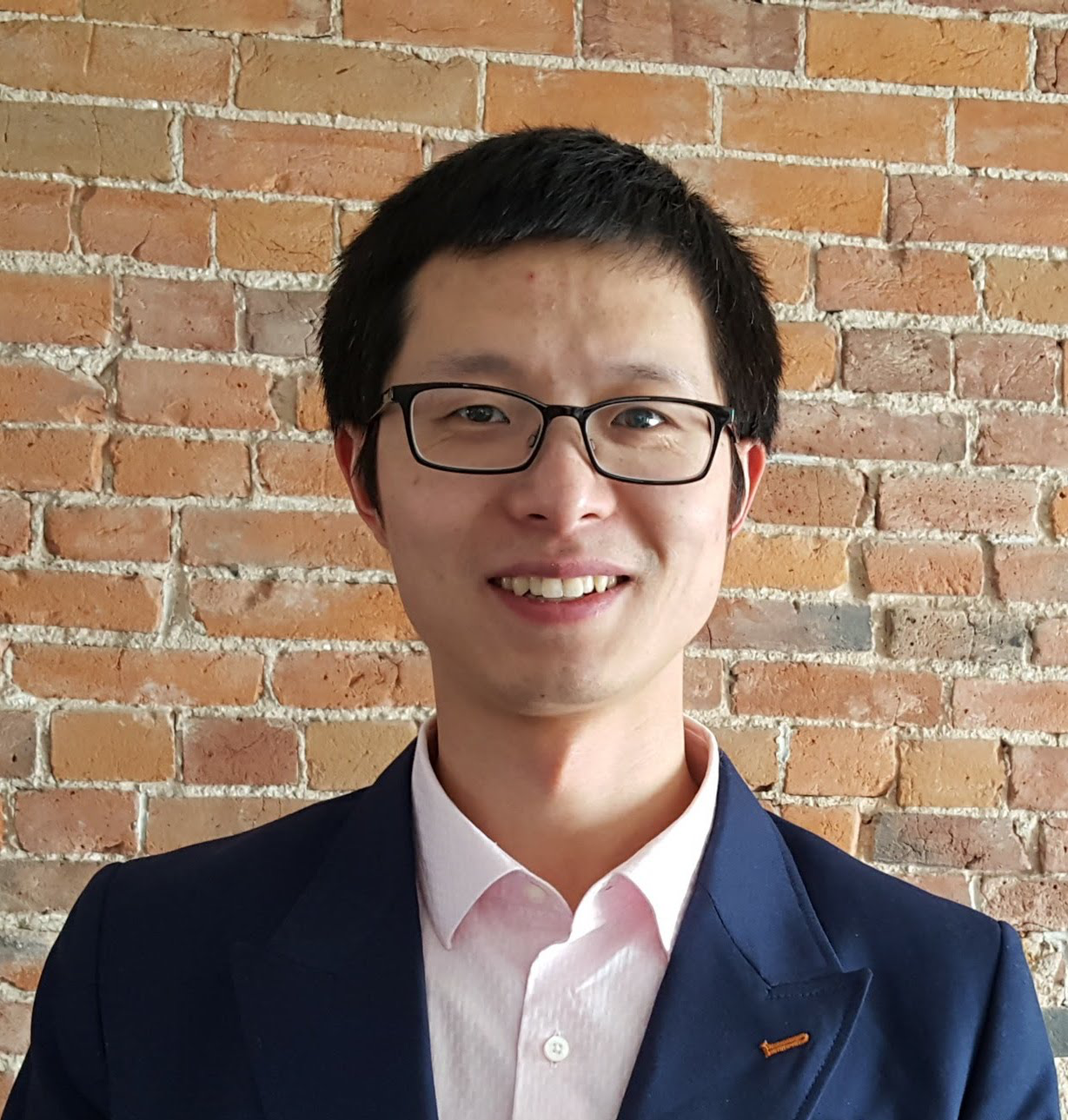}}]{Heng Li} 
Heng Li is an assistant professor in the Department of Computer and Software Engineering at Polytechnique Montreal, where he leads the MOOSE lab (moose.polymtl.ca). He holds a PhD in Computing from Queen’s University (Canada), M.Sc. from Fudan University (China), and B.Eng. from Sun Yat-sen University (China). Prior to his academic career, he worked in the industry for years as a software engineer at Synopsys and as a software performance engineer at BlackBerry. His and his students' research aims to address the practical challenges in software monitoring, software quality engineering, intelligent operations of software systems, and quality engineering of machine learning applications.
\end{IEEEbiography}








\end{document}

%% file: Text/abstract.tex
\begin{abstract}
Machine learning (ML), including deep learning, has recently gained tremendous popularity 
in a wide range of applications. 
However, like traditional software, ML applications are not immune to the bugs that result from programming errors. 
Explicit programming errors usually manifest through error messages and stack traces. These stack traces describe the chain of function calls that lead to an anomalous situation, or exception. Indeed, these exceptions may cross the entire software stack (including applications and libraries). Thus, studying the patterns in stack traces can help practitioners and researchers understand the causes of exceptions in ML applications and the challenges faced by ML developers.
To that end, 
 we mine Stack Overflow (SO) and study 11,449 stack traces related to seven popular Python ML libraries. 
First, we observe that ML questions that contain stack traces gain more popularity than questions without stack traces; however, they are less likely to get accepted answers.
Second, we observe that recurrent patterns exists in ML stack traces, even across different ML libraries, with a small portion of patterns covering many stack traces.
Third, we derive five high-level categories and 25 low-level types from the stack trace patterns: most patterns are related to \textit{python basic syntax}, \textit{model training}, \textit{parallelization}, \textit{data transformation}, and \textit{subprocess invocation}. Furthermore, the patterns related to \textit{subprocess invocation}, \textit{external module execution}, and \textit{remote API call} are among the least likely to get accepted answers on SO.  
Our findings provide insights for researchers, ML library providers, and ML application developers to improve the quality of ML libraries and their applications.


\end{abstract}

\begin{IEEEkeywords}
Machine learning applications, stack traces, Stack Overflow, empirical software engineering.
\end{IEEEkeywords}

%% file: Text/introduction.tex
\section{Introduction}
\label{sec:introduction}



\IEEEPARstart{T}{he} popularity of machine learning (ML) (including deep learning) applications has grown rapidly in recent years~\cite{stt6, ref8}.
Indeed, a $2020$ Deloitte survey of $2,737$ IT companies from nine countries found that modern 
machine learning technologies 
were being used by more than $50\%$ of these companies, and that $95\%$ of the respondents were planning to use them within the following year~\cite{deloitte}. Furthermore, in $2022$, The NewVantage Partners' \textit{Data and AI Leadership Executive Survey}  demonstrated that $91.7\%$ of organizations were expanding their investment in data and AI activities, at the core of which is ML~\cite{ref8}. 

\nd The growing trend in ML adoption delivers new challenges for ML developers (i.e., software developers and data scientists)
~\cite{islam2019comprehensive, zhang2019empirical, alshangiti2019developing, hamidi2021towards}. 
When ML developers face issues in their development process, they can turn to technical question and answer (Q\&A) forums such as Stack Overflow (SO) to answer their questions~\cite{alshangiti2019developing, bangash2019developers, islam2019developers, hamidi2021towards, zhang2019empirical}.
In this work, we analyze such questions to understand the stack traces that developers face when they develop ML programs. We target ML-related questions on SO because, with over $22$ million questions, $33$ million answers, and roughly $18$ million users, SO is the largest technical Q\&A forum~\cite{st2020new}. 
A $2021$ survey on SO~\cite{stSr2021} has shown that $80$ percent of respondents visit the forum weekly, and $50$ percent visit it daily. We, therefore, leverage SO to obtain data on exceptions that developers face when they develop ML programs because it is both popular and ingrained in developer work habits.

Prior works have studied the issues and challenges of developing machine learning applications~\cite{islam2019comprehensive, zhang2019empirical, alshangiti2019developing, hamidi2021towards,sun2017empirical,bangash2019developers}.
Indeed, it has been shown that the root causes of bugs and ML library challenges can be uncovered through a combination of bug reports and ML posts from  GitHub and SO.
However, the stack traces of ML applications remain unstudied. 

\nd A stack trace reports the chain of function calls at play when an anomalous situation, or exception, occurs. At the time of the exception, these function calls are either under execution or waiting for other function calls to be completed. 
The stack traces provide clues for understanding the errors that lead to exceptions (i.e., what chain of function calls lead to the exception). 
Therefore, in this work, we study the stack traces of ML applications to understand what the causes of exceptions in ML applications. 
Our work can provide a complementary perspective to existing works that study the issues and challenges of developing ML applications.

\nd In this work, we study \textasciitilde $11$K ML-related stack traces posted in SO questions, aiming to understand common patterns that cause exceptions in ML applications.  
To identify ML-related SO questions, we consider questions related to the use of seven popular Python ML libraries, including TensorFlow~\cite{tf}, Keras~\cite{ks}, Scikit-learn~\cite{sk}, PyTorch~\cite{pt}, NLTK~\cite{nl}, HuggingFace~\cite{hugh}, and Spark ML~\cite{sp}. 
We focus on the Python language as it is the dominant programming language for ML application development~\cite{pyref, pyref8}.
We perform quantitative and qualitative analyses to understand the characteristics of SO questions with stack traces and to analyze the patterns of stack traces therein. 
In particular, we structure our study and findings along the following three research questions (RQs):

\begin{itemize}
    \item \textbf{RQ1. What are the characteristics of ML-related posts with stack traces?} $12.6\%$ of posts related to our studied libraries include a stack trace. Although these posts receive more attention, they are less likely to have accepted answers, and take longer to be answered.
    \item \textbf{RQ2: What are the characteristics of the stack trace patterns in ML-related questions?} Most patterns are short, ($80\%$ of the patterns contain fewer than five calls), and $19.53\%$ are shared among at least two ML libraries. We also find that a small percentage of patterns ($20\%$) span a large number of questions (up to $85\%$).
    \item \textbf{RQ3: What are the categories of the stack trace patterns, and which categories are most challenging for developers?} Through a qualitative study, we identify five high-level categories and $25$ low-level types of stack trace patterns associated with ML development. We find that exceptions are often caused by misunderstandings or uncertainties related to ML API usage, data formats, or language constructs. Furthermore, exceptions related to external dependencies or manipulations of artifacts are less likely to receive timely community support.
\end{itemize}

\nd The main contributions of our work include:
\begin{itemize}
    \item This study is the first step towards understanding the stack traces of machine learning applications and their related issues posted on SO.
    \item We observe that questions with stack traces are less likely to get timely accepted answers. Forum moderators and researchers should pay particular attention to these questions to help developers.
    \item We observe patterns in stack traces of ML applications, and find that a small subset of them cover a wide range of questions, even across different libraries. Focusing on these prevalent patterns could produce impactful improvements.
    
    \item We provide a hierarchical taxonomy of the stack trace patterns of ML applications and recommend actions to improve ML applications and their usage. 
    \item Our observations identify the most challenging pattern types and provide critical insight for future research.
    
\end{itemize}

\nd The data and scripts used to produce this work are shared in a replication package~\cite{replication:Mining:2022}. 

%% file: Text/case_study_setup.tex
\section{Experiment setup}
\label{sec:approach}

\begin{figure}[ht]
    \center    
    \includegraphics[width=\linewidth]{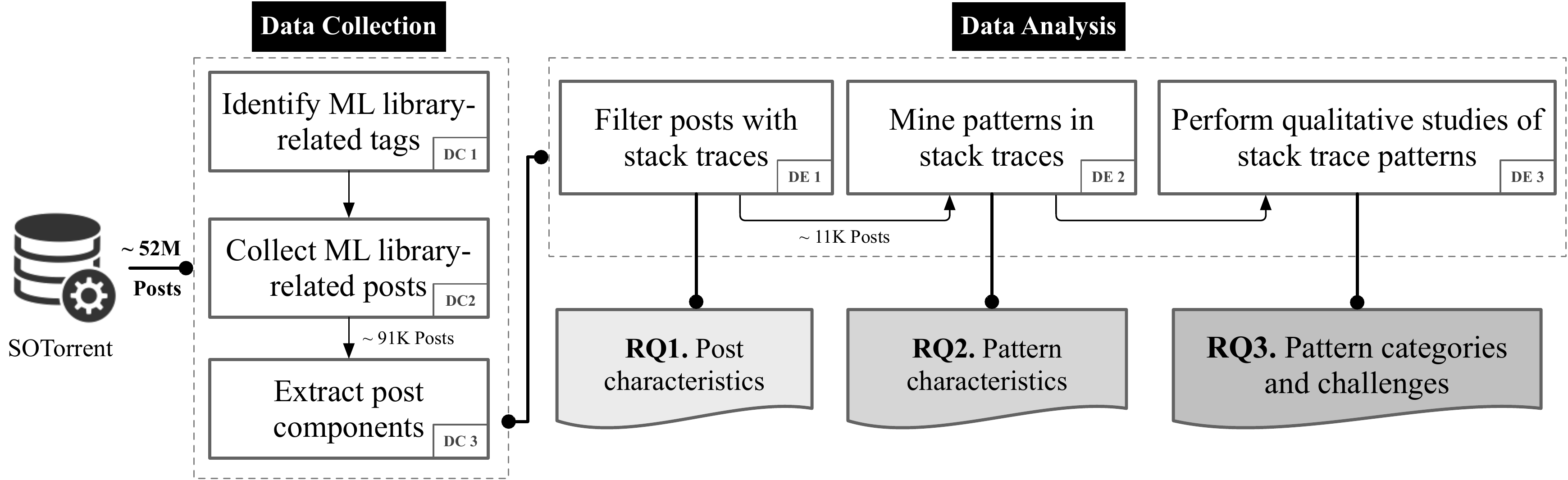}
    \vspace{-4mm}
    \caption{An overview of our study design}
    \label{fig:overview_2}
    \vspace{-2mm}
\end{figure}

Figure~\ref{fig:overview_2} gives an overview of our approach.
We first extract ML-related posts (which contain questions and answers) from SO. Then, we extract stack traces from the posts (through regular expressions) and use pattern mining techniques to discover patterns in the stack traces. The extracted stack traces and their patterns are used to answer our research questions. In the rest of this section, we elaborate on each of these steps to precisely capture each step of our experiment setup. The detailed analysis of each research question is presented in Section~\ref{sec:results}.


\subsection{\textbf{Subject Data}} 
We use SO posts as our primary data source to study the stack traces of ML applications. Specifically, we use SOTorrent~\cite{stt9}, an open dataset, to access the set of SO posts which contains questions, answers, and their metadata (e.g., question tags, view counts, scores, question and answer creation time, and whether an answer is accepted or not, etc). 
We use the last released version of SOTorrent, released on $2020$-$12$-$31$, and use 
Google BigQuery to access the dataset.
The dataset contains the entire version history of all SO posts in the official SO data dump from its first post on July $31$, $2008$, until $2020$-$12$-$31$.


\subsection{\textbf{Data Collection}}
\subsection*{\textbf{[DC1] Identify ML library-related tags}}
We use SO question tags to identify ML-related posts.
Using these tags, we focus on the questions related to the Python programming language and our selected ML libraries (e.g., TensorFlow). We choose the Python programming language because it is the most popular programming language for data scientists~\cite{pypop2, stSr2021} and involves libraries that are popular in the ML field~\cite{comp, ref22, tax}. To identify the tags related to Python ML libraries, we first define our studied libraries.

\nd \textbf{Defining studied ML libraries.} 
We consider SO questions related to seven popular ML libraries: \textit{TensorFlow}~\cite{tf}, \textit{PyTorch}~\cite{pt}, \textit{Scikit-learn}~\cite{sk}, \textit{Keras}~\cite{ks}, \textit{NLTK}~\cite{nl}, \textit{HuggingFace}~\cite{hugh}, and \textit{Spark ML}~\cite{sp}.
These popular and state-of-the-art libraries provide ready-made algorithms that allow ML developers to efficiently apply ML-based solutions in their applications. 
We select these seven ML libraries based on (1) their popularity metrics on GitHub~\cite{gitpop1, gitpop2} (e.g., the number of forks, stars, and releases), as well as (2) their usage in previous studies~\cite{comp, ref22, tax, stSr2021}.

\nd \textbf{Identifying a tag-set.} We leverage  StackExchange\footnote{\url{https://data.stackexchange.com/stackoverflow/queries}} and SO tags\footnote{\url{https://stackoverflow.com/tags}} to find and extract tag names related to ML libraries. The StackExchange query space provides an environment to find all of the existing tags that match a specified pattern through the \texttt{LIKE} operator. For example, to find the tags related to the TensorFlow library, we use the query \texttt{TagName LIKE '\%tensorflow\%' THEN 'tensorflow'}. This finds all of the versions and subcategories related to TensorFlow like \textit{tensorflow-lite} and \textit{tensorflow2.x}. We use a combination of queries to obtain a relevant tag-set for each of our studied ML libraries. Our replication package~\cite{replication:Mining:2022} contains our queries and our complete tag-set. 

\subsection*{\textbf{[DC2] Collect ML library-related posts}}
\nd To extract posts based on our tag-set, we use the Google Cloud Platform service (BigQuery). We export all post answers and ML-related questions using two  queries. We use the output of these queries to extract the specific part of the metadata that we need to answer our research questions. 

\subsection*{\textbf{[DC3]  Extract post components}}
\begin{figure}[ht]
    \center
    \includegraphics[width=\linewidth]{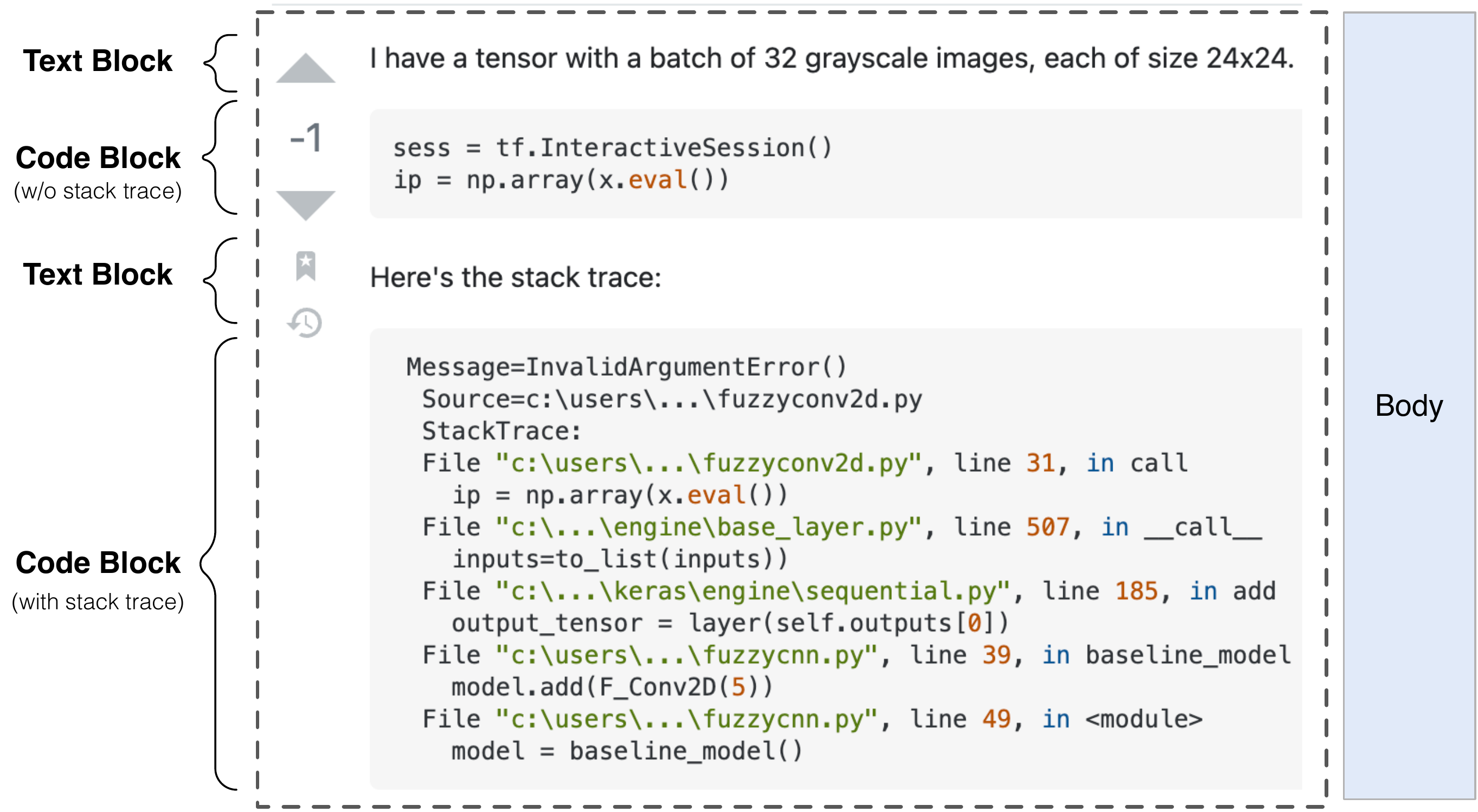}
    \caption{A truncated example of a SO question. (ID=52253378)}
    \label{fig:post_exm}
\end{figure}

\nd 
Posts on SO can be composed of text, images, code snippets, and more. For simplicity, as shown in Figure~\ref{fig:post_exm}, we summarize each post's body into text and code blocks. \textit{Text Blocks} contain the questioner's textual descriptions, and \textit{Code Blocks} are composed of code snippets or reports such as stack traces. In this study, we focus on posts that contain \textit{Code Blocks} with stack traces. Although SO often uses the (\texttt{<pre><code>}) tag to distinguish the \textit{Code Blocks}, there are cases where \textit{Code Blocks} obey other tags. Thus, we apply a lightweight heuristic approach that uses multiple regular expressions to identify \textit{Code Blocks} in the SO posts and automatically extract all \textit{Code Blocks} and \textit{Text Blocks}. Our heuristics are provided as scripts in our replication package.

\subsection{\textbf{Data Analysis}}

\subsection*{\textbf{[DE 1] Filter posts with stack traces}}
\nd Our research focuses stack traces. Therefore we wish to distinguish between questions with \textit{Code Blocks} with stack traces and other types of \textit{Code Blocks} such as code snippets. Regardless of operating system (OS) type, all stack traces contain OS paths. We use this knowledge to extract stack traces automatically through regular expressions.

\subsection*{\textbf{[DE 2] Mine patterns in stack traces}}
\begin{figure}[ht]
    \centering
    \includegraphics[width=.45\textwidth]{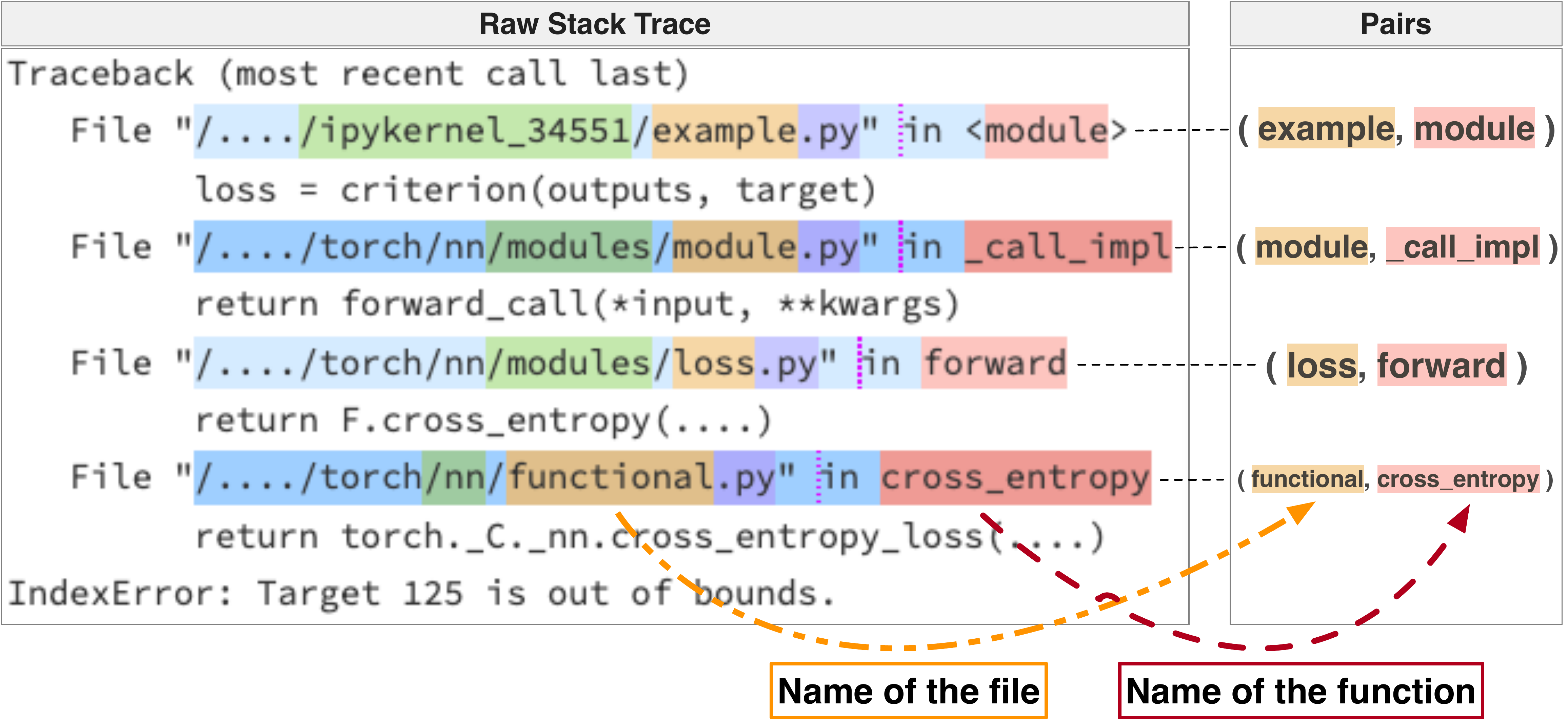}
    \caption{A truncated example of stack trace transformation. (ID=70716638)}
    \label{fig:create_pairs}
\end{figure}

\nd \textbf{Stack trace transformation.} The regular expressions used in DE1 provide us with two advantages: (1) we can differentiate between types of Code Blocks, and (2) we can capture the specific parts of each stack trace. Figure~\ref{fig:create_pairs} indicates a stack trace recognized using our regular expressions. Moreover, this figure shows how we convert each stack trace to a list of pairs. Each pair contains the file’s and function’s names, which come from each line in the stack trace. The combination of file and function names, the yellow and red colored words in Figure~\ref{fig:create_pairs}, gives us unique pairs that we use to feed our pattern mining procedure.

\nd \textbf{Stack trace mining.} In this paper we seek to uncover stack trace patterns and minimize the information lost due to the aggregation of different stack traces. For this purpose, we implement CC-Span~\cite{ccspan}, to mine closed contiguous sequential patterns. The algorithm considers two aspects: the items’ adjacency and the patterns’ closure. We use the unique pairs obtained from the previous step as input.

\subsection*{\textbf{[DE 3] Perform qualitative studies of stack trace patterns}} 
\nd We performed qualitative studies to identify categories of stack trace patterns, to better understand the types of stack trace-related issues faced by ML developers and their challenges. Our qualitative studies are detailed in Section~\ref{sec:RQ3}.


%% file: Text/results.tex
\section{Research questions and results} \label{sec:results}


\input{Text/RQ1}

\input{Text/RQ2}
\input{Text/RQ3}

%% file: Text/RQ1.tex

\begin{boxedSimple}
    {\textbf{(RQ1) What are the characteristics of ML-related posts with stack traces?}}
\end{boxedSimple}{}
\label{sec:RQ1}

\subsection*{\textbf{Motivation}}
When ML developers face challenges, they can post questions that contain stack traces on SO. These stack traces can indicate errors in their code, and communicate information about the challenges they face in their development process. Studying the characteristics of SO questions with stack traces can help ML developers and forum providers understand the profile of these questions and their level of community support. 
Therefore, this RQ proposes a quantitative analysis of the characteristics of ML-related SO questions with stack traces, and compares them statistically to SO questions without stack traces. 
This RQ also provides context for our next two RQs, where we aim to study the patterns in the stack traces of ML-related questions.

\subsection*{\textbf{Approach}}

In this RQ, we exclude post answers to concentrate on the characteristics of SO questions with stack traces and compare them to those without stack traces. In particular, we study the questions' lengths, the questions' popularity (e.g., view count), and how likely these questions are to receive answers that are accepted by the questioners.
\nd
\noindent\textbf{Studied characteristics of SO posts.} Below we describe the characteristic variables that we study to determine the status of ML-related stack traces on SO.

\begin{itemize}[label={}]
    \item \textbf{Question Length:} measures the number of words used in all \textit{Text Blocks} of a question.

    \item \textbf{Code Length:} measures the size of the \textit{Code Blocks} of a question. We count the number of lines in each \textit{Code Block} of a question. Our calculation ignores empty lines. 
    
    \item \textbf{Question Score:} a measure of popularity. The question score shows the difference between the number of upvotes and the number of downvotes for a question~\cite{score_ref}.
    
    \item \textbf{Comment:} a measure of popularity. A count of comments for a given question.
    
    \item \textbf{View:} a measure of popularity. SO counts the number of views for every page load\footnote{\url{https://meta.stackexchange.com/questions/36728/how-are-the-number-of-views-in-a-question-calculated}}. The total number of views for a given question. 
    
    \item \textbf{Answer Count:} measures the count of proposed answers for a given question.
    \item \textbf{First Answer:} measures the hours between a question's publication and when it receives its first answer.

    \item \textbf{Accepted Answer:} measures the hours between a question's publication and when it receives an accepted answer. Question posters can choose one answer that they believe is the best answer to their questions.
\end{itemize}


\nd We apply two statistical tests (i.e., Mann-Whitney U Test and Two-sample proportion Z Test) to compare these characteristics between questions with and without stack traces.

\nd \textbf{Mann–Whitney U Test.}
We apply this test~\cite{utest} to assess whether the data for ML-related stack traces can be considered statistically independent from questions w/o stack traces. We chose a nonparametric test because we do not know the distribution of our data a priory. 
Moreover, we execute the test at the $5\%$ significance level (i.e., $\alpha=0.05$). 



\nd \textbf{Two-sample proportion Z Test~\cite{ztest1, ztest2}.}
We apply this test in Table~\ref{tab:answer}, to compare the ratio of each ML library that gets accepted answers, for groups with and without stack traces. We use this statistical test because there is insufficient information to utilize the chi-squared distribution. We execute the test using a 5\% significance level.




\subsection*{\textbf{Results}}

\input{Table/tbl_post_num.tex}

\nd \textbf{An average of 12.6\% of ML-related questions provide stack traces.} 
 While the number of questions fluctuates based on the ML library, the percentage of questions with stack traces is relatively stable. As seen in Table~\ref{tab:post_cat_num}, TensorFlow ($13.6\%$) has the most questions with stack traces, and NLTK has the least ($10.0\%$).  This shows that a number of developers see an inherent value in adding stack traces to their SO questions. Furthermore, based on the stack trace paths, we find that most SO questions that contain stack traces come from Unix-based systems. This can be interpreted in two ways. Either most ML developers work and execute their code on Unix-based systems, or Unix-based developers are more likely to add a stack trace to their SO questions. Future work is needed to determine which is the case. However, in all cases, we find that stack traces are used in ML-related questions irrespective of the ML library in question. 

\input{Table/tbl_LOC.tex}

\nd \textbf{Questions with stack traces have shorter natural language descriptions and longer \textit{Code Blocks} compared to questions without stack traces.} Table~\ref{tab:length} displays the median text length and code length of questions with and without stack traces. Irrespective of library, questions with stack traces are less wordy (median $11$ fewer words on average) but use more lines of code (median $5$ more LOC on average) compared to questions without stack traces. 
Therefore, it appears that stack traces allow question posters to express their problem with fewer words by relying on the lines of code provided by the stack traces themselves.

\input{Table/tbl_community_support.tex}

\nd \textbf{Questions with stack traces gain more popularity (in terms of view counts and comment numbers) than questions without stack traces.} Table~\ref{tab:stat} compares the status of community support, including score, comment, view, and answer count among ML-related questions with and without stack traces against a baseline comprised of all Stack Overflow questions. While questions with stack traces garner lower scores than questions without stack traces, the median difference is small (\textasciitilde$1$). In terms of comment count and view count, questions with stack traces get more, or an equivalent number of comments and views compared to questions without stack traces and the SO baseline. Finally, we find that the median answer count amongst different ML libraries is the same in all groups, with a median answer count of $1$. Therefore, it appears that using stack traces in ML-library questions encourages comments and views.

\input{Table/tbl_answer.tex}

\nd \textbf{Questions with stack traces are less likely to get answers and take longer to get them when they do.}
We use the Two-sample proportion Z Test to compare the ratio of questions that receive an accepted answer with and without stack traces. Using Table~\ref{tab:answer}, we can see that questions with stack traces are less likely to receive an accepted answer. Indeed, for all statistically significant results, questions with stack traces take longer to obtain a first answer, and an accepted answer. While some libraries obtain answers faster than others, the trend holds across libraries, despite the overall answer speed. Indeed, Table~\ref{tab:answer} shows that questions with stack traces present unique characteristics that are distinct from questions without stack traces, even from questions that may contain other \textit{Code Blocks}, such as code snippets. 

\nd \textbf{Questions with stack traces rarely contain more than one stack trace.} Table~\ref{tab:thesame} presents the distribution of the stack traces on SO for each ML library. In all cases, questions rarely (at most $6\%$) contain more than one stack trace. 

\input{Table/tbl_post_vs_st.tex}

\begin{tcolorbox}[enhanced,attach boxed title to top left={yshift=-3mm,yshifttext=-1mm}, colback=black!5!white,colframe=black!75!black,colbacktitle=gray!80!black, title=Summary of RQ1,fonttitle=\bfseries,
  boxed title style={size=small,colframe=black!50!black} ]
  SO questions with stack traces are prevalent across all studied libraries, with $12.6\%$ of studied questions containing a stack trace. These questions generally use fewer words (a difference of $11$ words on average) and garner more comments and views than questions without stack traces. However, they are also less likely to have accepted answers, and take longer to be answered in the first place.
\end{tcolorbox}

%% file: Table/tbl_post_num.tex
\begin{table}[ht]
{
    \centering
    \small
    \renewcommand{\arraystretch}{1.2} 
    \caption{The Number of questions on SO for each studied ML library, including all questions, those that contain \textit{Code Blocks} and those that contain stack traces}
    \begin{adjustbox}{width=1\linewidth, center}
    \label{tab:post_cat_num}
    \begin{tabular}{l||r|wr{1cm}wl{1cm}|wr{1cm}wl{1cm}|cc}
        \multirow{2}*{\textbf{ML Libraries}} & \multirow{2}{*}{\textbf{\#~Ques.}} & \multicolumn{2}{c|}{\textbf{\#~Ques.}} &  \multicolumn{2}{c|}{\textbf{\#~Ques.}} & \multicolumn{2}{c}{\textbf{OS$^1$}} \\
         &  & \multicolumn{2}{c|}{with code block} &  \multicolumn{2}{c|}{with stack trace} & Unix-based & Windows \\
        \midrule
        \rowcolor{gray!15}
        TensorFlow & 39,690 &  32,368 &(81.6\%) & \textbf{5,436} &(13.6\%) & 69.1\% & 30.9\% \\
        Keras & 21,568 & 18,506 &(85.8\%) & \textbf{2,715} &(12.5\%) & 70.8\% & 29.2\% \\
        \rowcolor{gray!15}
        Scikit-learn & 18,133 & 14,795 &(81.6\%) & \textbf{2,009} &(11.0\%) & 62.0\% & 38.0\% \\
        PyTorch & 5,540 & 4,620 &(83.4\%) & \textbf{713} &(12.8\%) & 76.2\% & 23.8\% \\
        \rowcolor{gray!15}
        NLTK & 5,592 & 4,381 &(78.3\%) & \textbf{564} &(10.0\%) & 58.7\% & 41.3\% \\
        Hugging Face & 236 & 204 &(86.4\%) & \textbf{40} &(16.9\%) & 87.5\% & 12.5\% \\
        \rowcolor{gray!15}
        Spark ML & 141 & 126 &(89.4\%) & \textbf{16} &(11.3\%) & 93.8\% & 6.20\% \\
        \bottomrule
        \multicolumn{7}{l}{\small $^1$ OS stands for Operating System.} \\
    \end{tabular}
    \end{adjustbox}
}
\end{table}

%% file: Table/tbl_LOC.tex
\begin{table}[ht]
    \centering
    \small
    \renewcommand{\arraystretch}{1.2} 
    \caption{The median question and code lengths for the SO posts related to our studied ML libraries}
    \begin{adjustbox}{width=1\linewidth, center}
    \begin{tabular}{l|cclccl}
        \multicolumn{1}{c}{} & \multicolumn{3}{c}{\textbf{Ques. Length (Words)}} & \multicolumn{3}{c}{\textbf{Code Length (LOC)}} \\ \cmidrule(rl){2-4}\cmidrule(rl){5-7}
        \multicolumn{1}{c}{\textbf{ML Libraries}} & with stack trace & w/o$^1$ stack trace & \multicolumn{1}{c}{Sig.$^2$} & with stack trace & w/o stack trace & \multicolumn{1}{c}{Sig.} \\ \midrule
        \rowcolor{gray!15}
        \multicolumn{1}{l|}{TensorFlow} & 69.0 & 84.0 & *** & 25.0 & 19.0 & *** \\
        \multicolumn{1}{l|}{Keras} & 73.0 & 86.0 & *** & 24.0 & 21.0 & *** \\
        \rowcolor{gray!15}
        \multicolumn{1}{l|}{Scikit-learn} & 67.0 & 79.0 & *** & 18.0 & 14.0 & *** \\
        \multicolumn{1}{l|}{PyTorch} & 68.0 & 77.0 & *** & 19.0 & 17.0 & * \\
        \rowcolor{gray!15}
        \multicolumn{1}{l|}{NLTK} & 61.0 & 72.0 & *** & 15.0 & 11.0 & *** \\
        \multicolumn{1}{l|}{Hugging Face} & 67.5 & 90.0 & * & 18.0 & 14.0 & o \\
        \rowcolor{gray!15}
        \multicolumn{1}{l|}{Spark ML} & 83.5 & 80.0 & o & 43.0 & 16.0 & *** \\ \bottomrule
        \multicolumn{7}{l}{\small $^1$ w/o = without} \\
        \multicolumn{7}{l}{\small $^2$ Statistical significance of explanatory power according to Mann–Whitney U Test:} \\
        \multicolumn{7}{l}{\small o $ p \ge 0.05 $; * $ p < 0.05 $; ** $ p < 0.01 $; *** $ p < 0.001$}
    \end{tabular}
    \end{adjustbox}
\label{tab:length}
\end{table}

%% file: Table/tbl_community_support.tex
\begin{table*}[ht!]
    \centering
    \small
    \renewcommand{\arraystretch}{1.2} 
    \makegapedcells
    \caption{The median community support variables for SO posts of our studied ML libraries}
    \begin{adjustbox}{width=1\linewidth, center}
    \begin{tabular}{l|cclcclcclccl}
        \multicolumn{1}{c}{} & \multicolumn{3}{c}{\textbf{Ques. Score}} & \multicolumn{3}{c}{\textbf{Comment}} & \multicolumn{3}{c}{\textbf{View}} & \multicolumn{3}{c}{\textbf{Answer Count}} \\ \cmidrule(rl){2-4}\cmidrule(rl){5-7}\cmidrule(rl){8-10}\cmidrule(rl){11-13}
        \multicolumn{1}{c||}{\textbf{ML Libraries}} & with stack trace & w/o$^1$ stack trace & \multicolumn{1}{c}{Sig.$^2$} & with stack trace & w/o stack trace & \multicolumn{1}{c}{Sig.} & with stack trace & w/o stack trace & \multicolumn{1}{c}{Sig.} & with stack trace & w/o stack trace & \multicolumn{1}{c}{Sig.} \\ \midrule
        \rowcolor{gray!15}
        \multicolumn{1}{l||}{TensorFlow} & 0 & 1 & *** & 1 & 0 & *** & 324.0 & 202.0 & *** & 1 & 1 & **   \\
        \multicolumn{1}{l||}{Keras} & 1 & 0 & o & 1 & 1 & **  & 282.0 & 162.0 & *** & 1 & 1 & o \\
        \rowcolor{gray!15}
        \multicolumn{1}{l||}{Scikit-learn} & 1 & 1 & o & 2 & 1 & *** & 562.0 & 360.5 & *** & 1 & 1 & *** \\
        \multicolumn{1}{l||}{PyTorch} & 0 & 1 & ** & 0 & 0 & * & 345.0 & 168.0 & *** & 1 & 1 & o \\
        \rowcolor{gray!15}
        \multicolumn{1}{l||}{NLTK} & 1 & 1 & * & 2 & 1 & *** & 651.5 & 394.0 & *** & 1 & 1 & o \\
        \multicolumn{1}{l||}{Hugging Face} & 0 & 0 & o & 2 & 0 & ** & 134.0 & 131.5 & o & 1 & 1 & o \\
        \rowcolor{gray!15}
        \multicolumn{1}{l||}{Spark ML} & 1.5 & 1 & o & 0 & 0 & o & 597.5 & 651.0 & o & 1 & 1 & o \\ \midrule 
        \cline{2-3} \cline{5-6} \cline{8-9} \cline{11-12}
        \multicolumn{1}{l||}{\textbf{Baseline}} & \multicolumn{2}{c}{1} & - & \multicolumn{2}{c}{1} & - & \multicolumn{2}{c}{294} & - & \multicolumn{2}{c}{1} \\ 
        \cline{2-3} \cline{5-6} \cline{8-9} \cline{11-12}
        \bottomrule
        \multicolumn{7}{l}{\small $^1$ w/o = without} \\
        \multicolumn{7}{l}{\small $^2$ Statistical significance of explanatory power according to Mann–Whitney U Test:} \\
        \multicolumn{7}{l}{\small o $ p \ge 0.05 $; * $ p < 0.05 $; ** $ p < 0.01 $; *** $ p < 0.001$}
    \end{tabular}
    \end{adjustbox}
\label{tab:stat}
\end{table*}

%% file: Table/tbl_answer.tex
\begin{table*}[ht!]
    \centering
    \small
    \renewcommand{\arraystretch}{1.2} 
    \caption{The median time to answer for SO posts of our studied ML libraries}
    \begin{adjustbox}{width=1\linewidth, center}
    \begin{tabular}{l||rrlcrrlccl}
        \multicolumn{1}{c}{}  & \multicolumn{3}{c}{First Answer \textbf{(Hour)}} & \multicolumn{6}{c}{Accepted Answer \textbf{(Hour)}} \\ \cmidrule(l){2-4}\cmidrule(l){6-11}
        \multicolumn{1}{c}{\textbf{ML Libraries}}  & with stack trace & w/o$^1$ stack trace & \multicolumn{1}{c}{Sig.$^2$} & & with stack trace & w/o stack trace & \multicolumn{1}{c}{Sig.$^2$} & \multicolumn{1}{c}{ratio (with stack trace)} & \multicolumn{1}{c}{ratio (w/o stack trace)} & \multicolumn{1}{c}{Sig.$^3$} \\ \midrule
        \rowcolor{gray!15}
        TensorFlow & 5.82 & 1.88 & *** & & 3.96 & 2.40 & *** & 0.33 & 0.39  & *** \\
        Keras & 3.27 & 1.86 & *** & & 2.24 & 1.84 & ** & 0.36 & 0.39 & *** \\
        \rowcolor{gray!15}
        Scikit-learn & 2.97 & 1.53 & *** & & 2.50 & 1.56 & *** & 0.41 & 0.48 & *** \\
        PyTorch & 4.10 & 2.53 & *** & & 3.08 & 2.53 & o & 0.38 & 0.42 & *** \\
        \rowcolor{gray!15}
        NLTK & 2.39 & 1.14 & *** & & 2.91 & 1.16 & *** & 0.40 & 0.50 & *** \\
        Hugging Face & 21.24 & 18.48 & o & & 0.98 & 14.10 & o & 0.27 & 0.42 & o \\
        \rowcolor{gray!15}
        Spark ML & 2.40 & 2.43 & o & & 1.97 & 1.79 & o & 0.50 & 0.49 & o  \\ \bottomrule
        \multicolumn{9}{l}{\small $^1$ w/o = without} \\
        \multicolumn{9}{l}{\small $-$ Statistical significance of explanatory power according to Mann–Whitney U Test$^2$ and Two-sample proportion Z Test$^3$:} \\
        \multicolumn{7}{l}{\small o $ p \ge 0.05 $; * $ p < 0.05 $; ** $ p < 0.01 $; *** $ p < 0.001$}
    \end{tabular}
    \end{adjustbox}
\label{tab:answer}
\end{table*}

%% file: Table/tbl_post_vs_st.tex
\begin{table}[ht]
    \centering
    \small
    \renewcommand{\arraystretch}{1.2} 
    \caption{The distribution of stack traces on SO for our studied ML libraries}
    \begin{adjustbox}{width=1\linewidth, center}
    \begin{tabular}{lrrc}
    \\
        \multicolumn{1}{l}{\multirow{-2}{*}{\textbf{ML Libraries}}} & \multirow{-2}{2cm}{\centering \textbf{\#~Stack Trace \\ Instances}} & \multirow{-2}{3cm}{\centering \textbf{\#~Ques.}\\(with stack trace)} & \multicolumn{1}{l}{\multirow{-2}{*}{\textbf{Ratio}}} \\ 
        \hline
        \rowcolor{gray!15}
        \multicolumn{1}{l||}{TensorFlow} & 5,728 & 5,436 & 1.05 \\
        \multicolumn{1}{l||}{Keras} & 2,851 & 2,715 & 1.05 \\
        \rowcolor{gray!15}
        \multicolumn{1}{l||}{Scikit-learn} & 2,090 & 2,009 & 1.04 \\
        \multicolumn{1}{l||}{PyTorch} & 741 & 713 & 1.03 \\
        \rowcolor{gray!15}
        \multicolumn{1}{l||}{NLTK} & 585 & 564 & 1.03 \\
        \multicolumn{1}{l||}{Hugging Face} & 40 & 40 & 1.00 \\
        \rowcolor{gray!15}
        \multicolumn{1}{l||}{Spark ML} & 17 & 16 & 1.06 \\  
        \bottomrule
    \end{tabular}
    \end{adjustbox}
\label{tab:thesame}
\end{table}

%% file: Text/RQ2.tex

\begin{boxedSimple}{
    \textbf{(RQ2) What are the characteristics of the stack trace patterns in ML-related questions?}}
\end{boxedSimple}{}
\label{sec:RQ2}

\subsection*{\textbf{Motivation}}
\nd When a Python program faces an error or exception, the Python interpreter produces a stack trace that lists the function calls that lead up to the error. By examining the stack traces, we may identify patterns in the sequence of function calls that can help developers and researchers understand the root cause of the errors. 
In addition, finding frequent patterns among many stack traces may help us identify buggy functions and APIs for ML libraries.
Finally, stack trace patterns can also help us identify code structure or design problems that may contribute to errors.


\subsection*{\textbf{Approach}} 
To answer RQ$2$, we first seek to identify stack trace patterns among ML-related questions. Transforming stack traces into lists of pairs of file and function names (see Figure~\ref{fig:create_pairs}) allows us to create a list of all stack trace instances for each ML library. We use these lists as input to the CC-Span algorithm to uncover stack trace patterns. By grouping similar stack traces together and identifying patterns in the stack traces, we can gain a better understanding of the common issues and errors that occur when working with ML libraries.



\begin{figure}[ht]
    \centering
    \includegraphics[width=.45\textwidth]{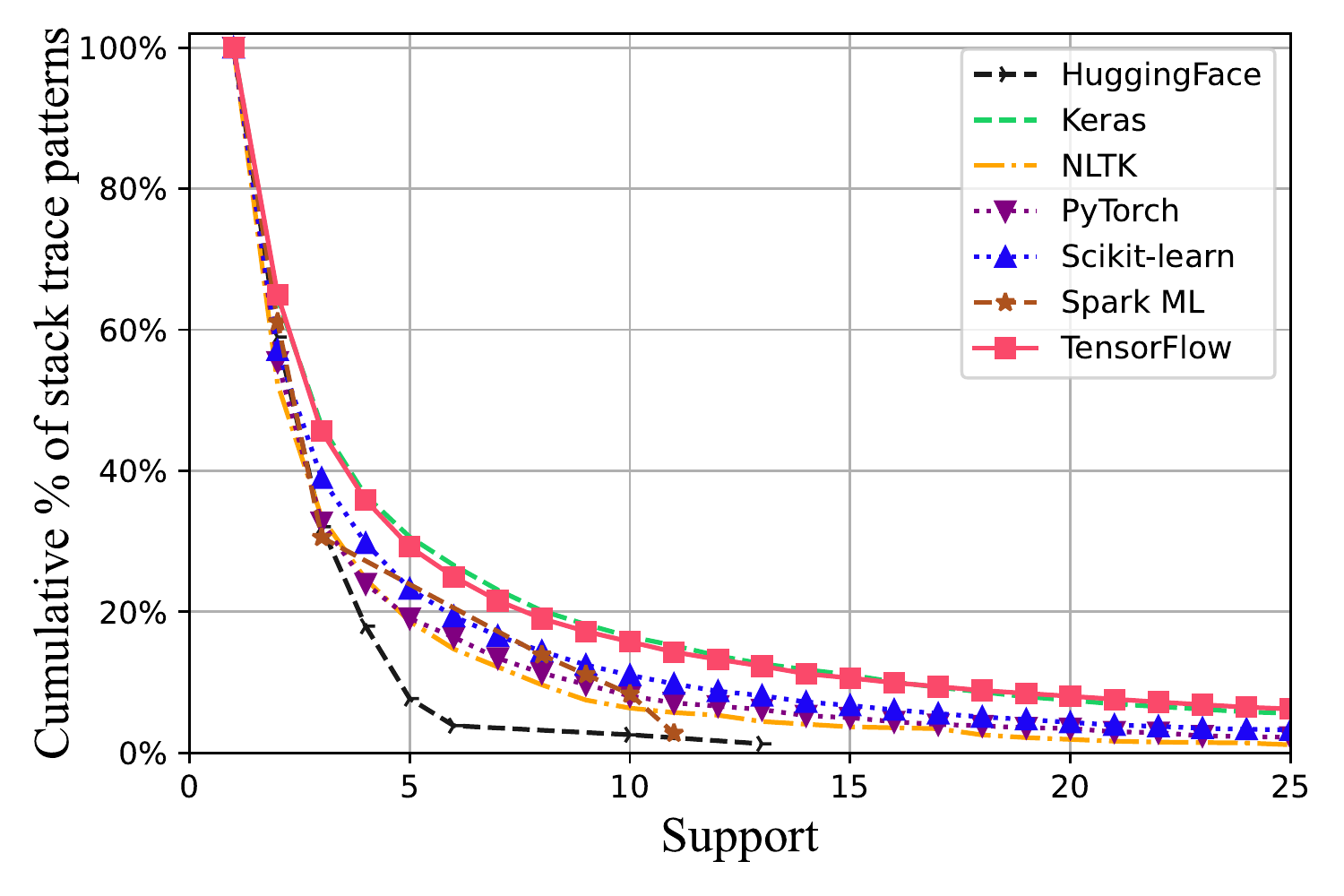}
    \caption[Short text]{The cumulative percentage of stack trace patterns for various supports $(support \geq 1)$}
    \label{fig:CC-Span-sup-all}
\end{figure}

\begin{figure}[ht]
    \centering
    \includegraphics[width=.45\textwidth]{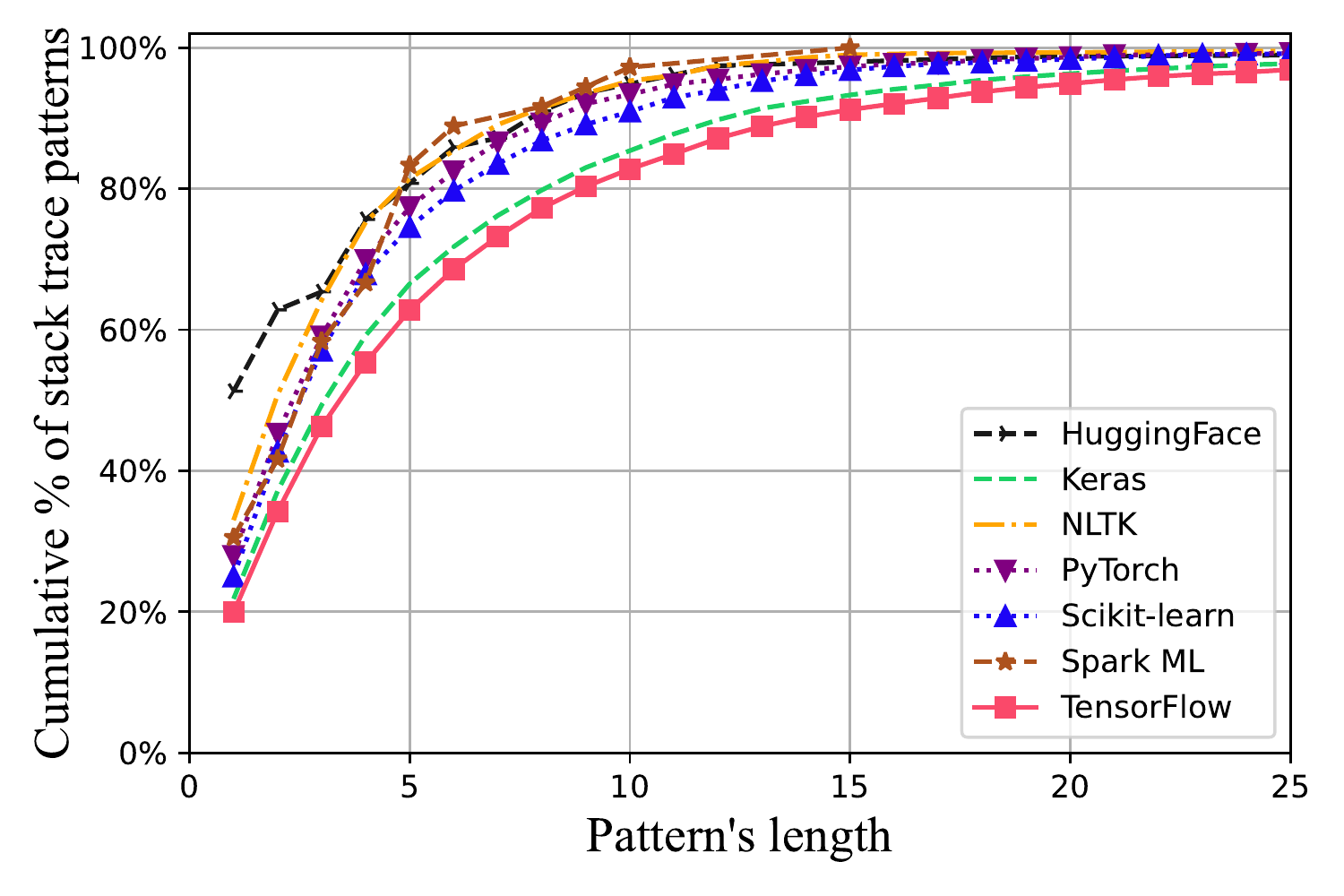}
    \caption[Short text]{The cumulative percentage of stack trace patterns vs. pattern’s length $(support \geq 1)$. 
    }
    \label{fig:CC-Span-length-all}
\end{figure}

\nd The CC-Span algorithm uses a support threshold as an input parameter. Figures~\ref{fig:CC-Span-sup-all} and \ref{fig:CC-Span-length-all} display the output and the cumulative percentage of stack trace patterns for various supports and pattern lengths. These plots show that by setting the $support$ = $2$, on average, we can enclose $83.88\%$ and $64.73\%$ of questions and stack trace instances in the stack trace patterns set. Thus, we set the support threshold to the smallest number ($2$) to reach the highest coverage. We ignore singleton instances ($support$ = $1$) because they are non-recurring and therefore not patterns. Table~\ref{tab:maintain_pattern} depicts the percentages of stack trace instances and questions that were maintained by using the support threshold value of $2$.



\input{Table/maintian_patterns.tex}

\subsection*{\textbf{Results}}

\noindent\textbf{Recurrent patterns are common in the stack traces of ML-related questions}.
Table~\ref{tab:maintain_pattern} reveals the percentage of stack trace instances and questions that fall into pattern sets with a support of at least $2$, grouped by different ML libraries. 
Indeed, a high proportion of stack trace instances clearly have patterns, and on average, we cover $83.88\%$ and $64.73\%$ of total questions and stack traces, respectively. Furthermore, recurrent patterns are common.

\begin{figure}[ht]
    \centering
    \includegraphics[width=0.45\textwidth]{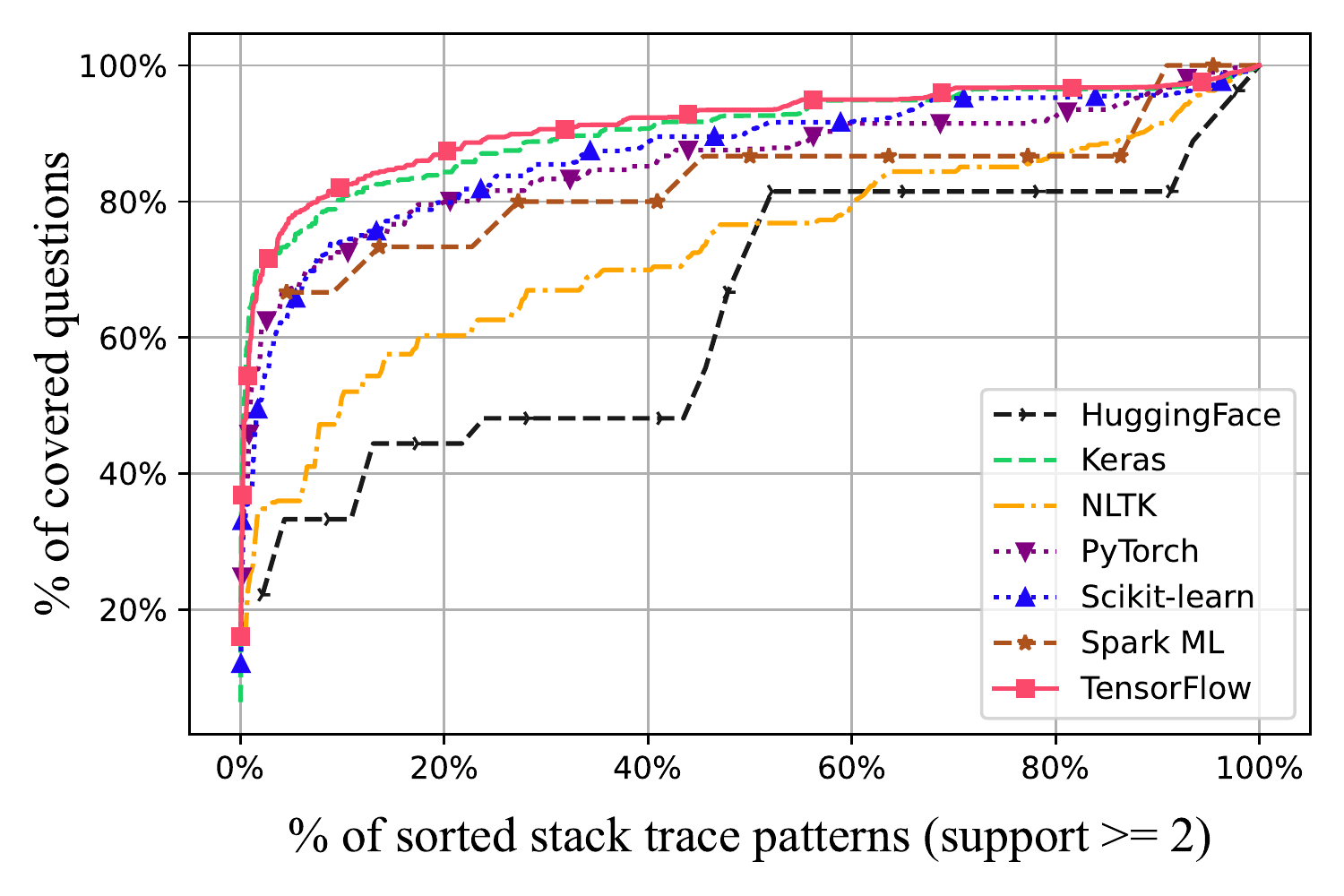}
    \caption[Short text]{The percentage of covered questions based on stack trace patterns (patterns are sorted based on their supports).}
    \label{fig:q_p_rq2.5}
\end{figure}

\noindent \textbf{A relatively small portion of patterns (20\%) cover a large number of questions
(75\% to 85\%).} Figure~\ref{fig:q_p_rq2.5} indicates the percentage of questions with stack traces that are covered by the percentage of stack trace patterns, for each ML library. 
For the majority of our studied libraries (i.e., not
HuggingFace and NLTK libraries), $20$ percent of significant patterns (high-support patterns) constitute a considerable proportion of questions ($75\%$ to $85\%$) for each ML library, and the rest ($80\%$) cover around $20\%$ of all questions. 
As shown in Figure~\ref{fig:CC-Span-sup-all}, for all ML libraries, many stack trace patterns ($20$\%-$65$\%) have low support (between $2$ and $4$); Less than $20\%$ of stack patterns have a support equal to or greater than 10. The detailed distribution of the support values of each ML library's patterns can be found in our replication package~\cite{replication:Mining:2022}.
Our results indicate the significance of studying these stack trace patterns as they cover the majority of the questions with stack traces.

\noindent\textbf{Most patterns are composed of few calls (short pattern length), with 80\% of patterns having a length shorter than or equal to 5.} Figure~\ref{fig:CC-Span-length-all} shows the percentage of stack traces found for different pattern lengths. Except for the Spark ML and HuggingFace libraries, $50\%$ of stack trace patterns have a length between $1$ and $3$, and few patterns (\textasciitilde $20\%$) are lengthy (more than $5$ calls). 
This may indicate that many of the issues and errors encountered when working with ML libraries can be traced back to a relatively small number of specific function calls or code blocks. This provides opportunities to study these short patterns and identify their root causes. Also, understanding these common patterns can be helpful for developers who are seeking answers to their problems, as it can help them identify error root causes and find solutions more quickly.



\begin{figure}[ht]
    \centering
    \includegraphics[width=.45\textwidth]{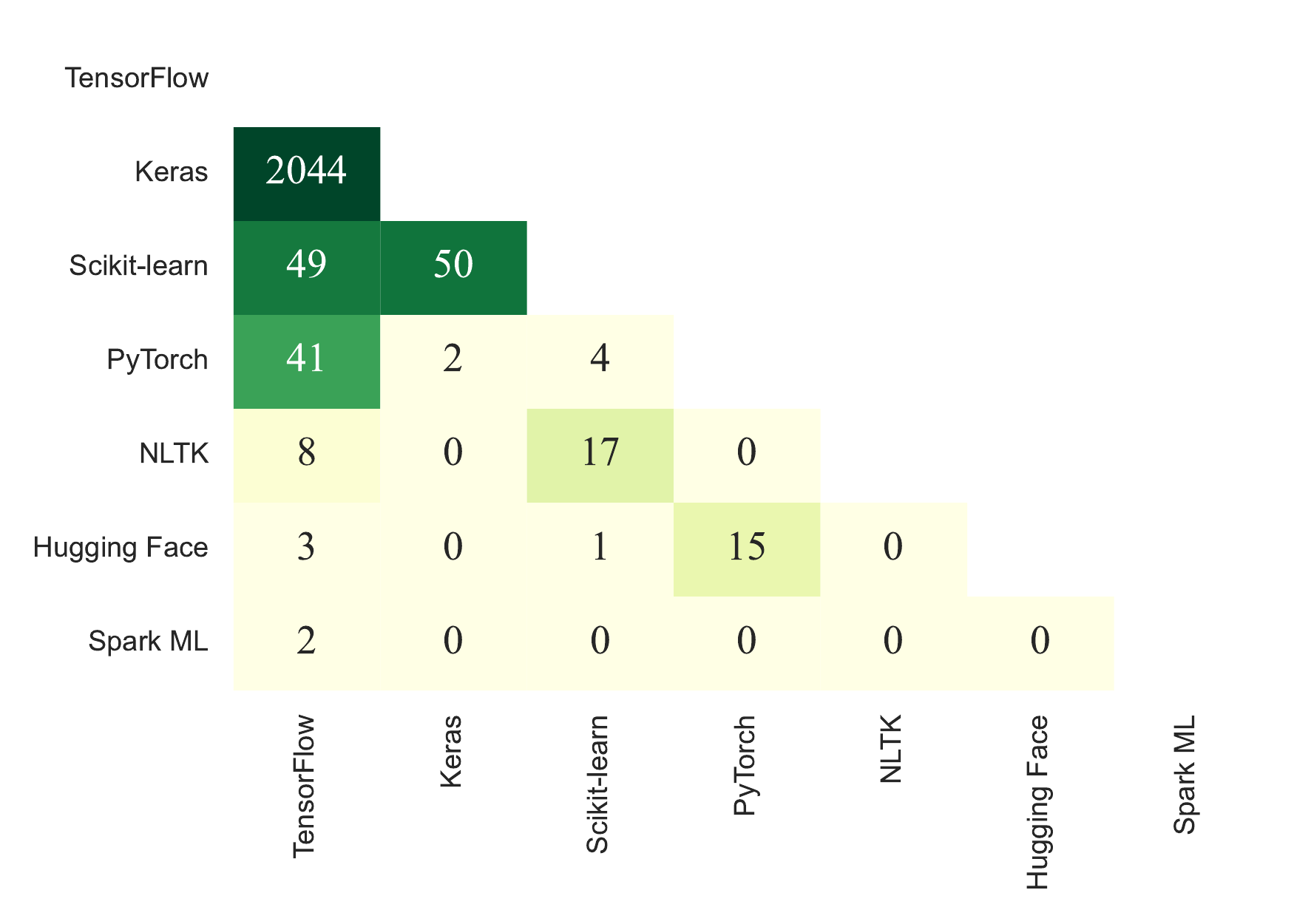}
    \caption[Short text]{Stack trace pattern correlation between ML-libraries}
    \label{fig:heat_map}
\end{figure}

\noindent\textbf{Stack trace patterns are shared across the questions of different ML libraries.}
Figure~\ref{fig:heat_map} and Table~\ref{tab:shared_pattern} give insight into the number of patterns shared between ML libraries. Figure~\ref{fig:heat_map} is a heatmap plot that uses a warm-to-cool color spectrum to demonstrate the correlation between pairs of ML libraries using their shared stack trace patterns. Table~\ref{tab:shared_pattern} shows how often libraries share patterns. As can be seen in Table~\ref{tab:shared_pattern}, few patterns are shared by many libraries. However, some pairs of libraries like TensorFlow and Keras have many ($2,044$) shared pattern. Indeed, TensorFlow shares patterns with many other ML libraries. 
This could indicate that these libraries encounter similar issues and errors. Analyzing shared stack trace patterns can be useful for identifying common problems when working with these ML libraries. By understanding these patterns, developers may anticipate and troubleshoot potential issues when working with these libraries, helping them find solutions more efficiently.


\input{Table/shared_patterns.tex}



\begin{tcolorbox}[enhanced,attach boxed title to top left={yshift=-3mm,yshifttext=-1mm},
  colback=black!5!white,colframe=black!75!black,colbacktitle=gray!80!black,
  title=Summary of RQ2,fonttitle=\bfseries,
  boxed title style={size=small,colframe=black!50!black}]
  Recurrent patterns are common in the stack traces of ML-related questions, and a relatively small portion of these patterns cover many questions. Most patterns have a short length, and some patterns are shared across the questions of different ML libraries.
\end{tcolorbox}

%% file: Table/maintian_patterns.tex
\begin{table}[ht]
    \centering
    \small
    \renewcommand{\arraystretch}{1.2} 
    \caption{What percent of questions and stack traces fall into the pattern set $(support \geq 2)$ 
    }
    \begin{adjustbox}{width=.8\linewidth, center}
    \begin{tabular}{lcc}
    \toprule
         \textbf{ML Libraries} & \textbf{Covered Ques. (\%)} & \textbf{Covered ST$^1$ (\%)} \\ 
        \hline
        \rowcolor{gray!15}
        \multicolumn{1}{l||}{TensorFlow} & 91.45 & 64.96 \\
        \multicolumn{1}{l||}{Keras} & 89.09 & 64.43 \\
        \rowcolor{gray!15}
        \multicolumn{1}{l||}{Scikit-learn} & 85.39 & 57.10 \\
        \multicolumn{1}{l||}{PyTorch} & 82.44 & 55.42 \\
        \rowcolor{gray!15}
        \multicolumn{1}{l||}{NLTK} & 77.58 & 52.28 \\
        \multicolumn{1}{l||}{Hugging Face} & 67.50 & 58.97 \\
        \rowcolor{gray!15}
        \multicolumn{1}{l||}{Spark ML} & 93.75 & 100 \\  
        \bottomrule
        \multicolumn{1}{l||}{\textbf{Overall}} & \textbf{83.88} & \textbf{64.73} \\
        \bottomrule
        \multicolumn{3}{l}{\small $^1$ ST stands for Stack Trace.}
    \end{tabular}
    \end{adjustbox}
\label{tab:maintain_pattern}
\end{table}


%% file: Table/shared_patterns.tex

\begin{table}[ht]
    \centering
    \small
    \renewcommand{\arraystretch}{1.2} 
    \caption{The number of shared stack trace patterns amongst various ML libraries $(support \geq 2)$}
    \begin{adjustbox}{width=.8\linewidth, center}
    \begin{tabular}{lrrrrr}
    \toprule
    \textbf{Shared Libraries (\#)} & 2 & 3 & 4 & 5 & 6 \\
    \textbf{Stack trace Patterns (\#)} & 2,236 & 145 & 47 & 20 & 1 \\
    \bottomrule
    \end{tabular}
    \end{adjustbox}
\label{tab:shared_pattern}
\end{table}

%% file: Text/RQ3.tex
\begin{boxedSimple}
    \textbf{(RQ3) What are the categories of stack trace patterns, and which categories are most challenging for developers?}
\end{boxedSimple}{}
\label{sec:RQ3}

\subsection*{\textbf{Motivation}}

In the previous RQs, we observe that stack traces are prevalent in ML-related questions on SO (RQ$1$), and that these stack traces share common patterns (RQ$2$). 
To have a deeper understanding of the stack trace patterns, in this RQ, we perform a qualitative study to categorize them. We also analyze which categories of patterns receive less community support on SO.
Our results can provide insights into the causes of exceptions in ML applications and shed light on future work to help practitioners overcome their challenges.

\subsection*{\textbf{Approach}} 
We perform our qualitative study in three rounds (i.e., Round $1$, $2$, and $3$). In Round $1$, we employ the stack trace patterns shared among ML libraries to cover a diversity of ML libraries. In Round $2$, to consider the patterns’ impact and sample from all captured patterns, we purposefully sample the stack trace patterns to assign a higher sampling probability to the patterns with higher support. In 
the last round, Round $3$, we derive stack trace categories. We explain each round in detail in the following subsections. 


\nd \textbf{[Round 1 of our qualitative study]}

\nd 
In the first step, we look for shared patterns among ML-related libraries. Using Table~\ref{tab:shared_pattern}, we choose patterns that are shared among $3$, $4$, $5$, and $6$ libraries ($213$ stack trace patterns). To understand the pattern types, we manually analyze these $213$ patterns. We explain our first qualitative analysis process in detail below. 

\nd \textbf{Qualitative analysis of shared patterns.} We used grounded theory~\cite{open_coding_1, open_coding_2, open_coding_3, gt} to identify stack trace pattern categories. Grounded theory has three phases: open, axial, and selective coding and is appropriate for concluding a high-level abstraction from low-level definitions~\cite{gt}. 

For each pattern instance, we randomly fetch five associated posts. We examine the file's path, the function's name, and the error line in each stack trace. We use this information to understand the reason behind the error report. 

We assign one type (label) for each pattern instance. If a stack trace pattern could belong to multiple types, we identify and choose its primary type. All three authors of the paper jointly conduct the coding process by following the five phases listed below:

\begin{steps}
    \item \textbf{Phase-1 coding.} We randomly shuffle all $213$ patterns and divide them into three parts. We distribute the first third of all patterns amongst three coders. Each coder labels (i.e., assigns a type to) them separately.
    
    \item \textbf{Discussion after phase-1 coding.} We discuss and analyze our resulting types and reach an agreement in order to obtain the same types for all coders.
    
    \item \textbf{Revisiting phase-1 coding.} According to the result of our discussion, we revise the phase-1 coding results to obtain consistent types.
    
    \item \textbf{Phase-2 coding.} In this phase, we again divide the remaining portion of the patterns (i.e., two-thirds) into three parts and assign each part to two different coders. Coders can still add new types during this phase. 
    
    \item \textbf{Discussion after phase-2 coding and resolving disagreements.} We discuss and analyze types updated in phase-2 and reach a consensus. We finalize the types after this step. During this final evaluation, if we find a disagreement between two coders, the third coder resolves the disagreement for the particular pattern. 
\end{steps}

\nd \textbf{Measuring the reliability of Round-1 results.} We calculate inter-rater reliability to measure the validity of our qualitative study results. The results of our qualitative studies are reliable if raters (or coders) reach a certain level of agreement on their labeling procedure.

\nd \textbf{Krippendorff's $\alpha$~\cite{alpha_range, k10}.} We use $\alpha$ to measure the agreement achieved after the coding process. $\alpha$ supports any number of coders as well as a standard coefficient for estimating inter-rater reliability. $\alpha$ is given by:
\begin{align} 
    \mathbf \alpha = 1- \frac{D_0}{D_e}
\end{align}

\nd where $D_0$ is the disagreement observed between multiple coders and $D_e$ is the disagreement expected by chance. The $\alpha$ value ranges between $0$ and $1$, where $0$ is the complete absence of agreement and $1$ indicates perfect agreement between coders~\cite{alpha_range}. 
We obtain a Krippendorff’s $\alpha$ of $0.885$ for this qualitative study, which indicates a reliable agreement (i.e., common understanding) between the coders. 

\nd \textbf{[Round 2 of our qualitative study]}

\nd 
In our second qualitative study, we expand our investigations to all stack trace patterns instead of just shared ones (as done in Round 1). However, we obtain a total of $11,449$ stack trace patterns, too many for a manual investigation. To select a reasonable number of patterns, we focus our sampling on ``critical patterns'' by giving patterns with a higher support a proportionally higher chance to be sampled. 

\nd We use equation~\ref{eq:ceil} to calculate how many patterns to sample, where $A_i$ is the set of all of the stack trace patterns for library $i$. Meanwhile, $n$ represents the total number of sampled stack trace patterns, and $n_i$ represents the number of sampled patterns for library $i$.
\begin{align} 
    \label{eq:ceil}
    \mathbf ceil\left \lceil \frac{len(A_i) \times n}{\sum_{i=1}^{7}len(A_i)} \right \rceil = n_i
\end{align}

To ensure that we choose at least one pattern for each ML library, we use Eq.~\ref{eq:ceil} with $i$ = $1$, and $n_1$ = $1$, to force a pattern to be selected from the library with the least patterns (i.e., Spark ML with $22$ patterns), this implies $len(A_1)$ = $22$. Furthermore, we have a total of 11,449 patterns, which implies that $\sum_{i=1}^{7}len(A_i)$ = $11,449$. Using these numbers we can calculate $n$. Because we use $ceil$ to round our result, there is some flexibility in the result. We choose to use the highest possible result. This yields a total of $782$ patterns (i.e., $n$). We then use this to calculate sample sizes for other libraries. Details are available in our replication package. 

\nd Using Eq.~\ref{eq:ceil} and the types identified in Round $1$, we sample $782$ patterns for the second part of our qualitative study.

\nd \textbf{Qualitative analysis of selective patterns.} We perform a hybrid coding approach and obey the same structure as the previous Round (i.e., randomly selecting five posts, analyzing the posts containing the patterns, etc). We use types from Round $1$ and extend them in Round $2$. 
As the coders have established a reliable common understanding of the labels in Round $1$, the first author of the paper is the primary coder in Round $2$. The other two authors joined the first author to help resolve uncertain labels and discuss the coding results.
The following steps are listed below:
\begin{steps}
    \item \textbf{Phase-1 coding.} One of the coders (the first author) codes all $782$ patterns based on the types which come from Round $1$. New types are added if they do not exist in the past coding process. 
    
    \item \textbf{Discussion and Revisiting after Step-1 coding.} We hold a meeting to analyze the first step's results. This meeting lasts about two hours, intending to reach an agreement. Also, during this meeting, all coders talk about allocating final types for those patterns and types that are difficult to code for the first coder. 
    
    \end{steps}

\nd \textbf{[Round 3 of our qualitative study]}

\nd \textbf{Defining stack trace pattern categories.}
We use Axial coding~\footnote{\url{https://delvetool.com/blog/axialcoding}} to find an answer for the second part of RQ$3$. Axial coding is a method of analysis used in qualitative research to recognize patterns and relationships within data. It involves breaking the data into smaller units and assigning them codes or labels. To that end, all three coders hold a discussion meeting and define the categories. In this meeting, we print the stack trace pattern types onto cards and classify them into categories. Then we merge and split certain stacks of cards through a short discussion as the need arises. The resulting categories are discussed in the section below.

\subsection*{\textbf{Results}}




\noindent\textbf{We identified 25 stack trace pattern types across five high-level categories, including model-related patterns, data-related patterns, Python language syntax-related patterns, external dependence-related patterns, and multi-process-related patterns.} 
In total, we manually analyzed $995$ stack trace patterns. This sample covers $8.2\%$ of all stack traces in our dataset. Table~\ref{tab:labels} shows the open coding results after all three rounds of our qualitative studies, including the pattern category, pattern type, and frequency of the pattern types in Round $1$ and $2$, separately. The frequency column (\textit{Freq.}) shows the total number of patterns categorized into a specific pattern type, the frequency of the pattern type in each round, and the percentage of patterns of that type. The table is sorted based on the number of patterns for each pattern type. 
Overall, we categorize the $995$ manually analyzed stack trace patterns into the $25$ distinct pattern types defined in the description column of Table~\ref{tab:labels}. An example of each pattern type can be found in the Example Post column. 
Some ($9$) patterns were not categorized 
during the coding process. These patterns follow uncommon stack trace structures; we therefore deliberately ignore them and do not show them in the result table. 

Although the number of patterns and the sampling approaches used in both rounds of our qualitative study are different, as presented in Table~\ref{tab:labels}, there are few differences in the pattern types found in both rounds. Indeed, in the second round, we only add four new pattern types, including \textit{Package installation}, \textit{Model conversion}, \textit{Data loading}, and \textit{Data conversion}. Finally, \textit{Model copy} is the only pattern that exists in Round $1$ but not in Round $2$. In total, the share of new pattern type instances in Round $2$ is less than $2.5\%$ of our sample. 
We therefore believe that we reached type saturation and identified representative patterns. The identified stack trace pattern categories are discussed below.

\input{Table/tbl_labels.tex}


\nd \textit{\textbf{[MOD] Model-related patterns}}. Pattern types classified into this category are related to the models of ML algorithms. Specifically, the model-related pattern types include model training, hyperparameter tuning, model validation, model saving/loading, model conversion,  and model copying. The first three types are related to building and validating the models, while the latter types are related to manipulating the model artifacts.  
\begin{minipage}{\linewidth}
\begin{lstlisting}[language=python, label=mod_example, caption={\unexpanded{[MOD]} Model-related patterns: truncated example (ID=$57842734$). The failure is related to a dimension mismatch in the model training process.}, belowskip=-0.1 \baselineskip]
    Traceback (most recent call last)
    |\lcolorbox{yellow}{/.../def\_function in \_\_call\_\_}|
    |\lcolorbox{yellow}{/.../training\_v2.py in fit}|
    |\lcolorbox{yellow}{/.../training\_v2.py in run\_one\_epoch}|
    |\lcolorbox{yellow}{/.../training\_v2\_utils.py in execution\_function}|
    |\lcolorbox{yellow}{/.../def\_function.py in \_\_call\_\_}|
    /.../function.py in __call__
    /.../function.py in _filtered_call
    /.../function.py in _call_flat
    /.../function.py in call
    /.../execute.py in quick_execute
    /.../six.py in raise_from
\end{lstlisting}
\end{minipage}
\nd \textbf{Model-related exceptions are often caused by misunderstanding the input requirements of the model APIs or failing to meet such requirements.}
ML algorithms are often implemented as libraries (e.g., TensorFlow) and used as black boxes by developers~\cite{castelvecchi2016can}.
The behaviors of the ML algorithms are abstracted by the APIs provided by the libraries. 
However, developers may misunderstand the behaviors of the APIs(e.g., providing incorrect arguments or input data).
Such misunderstanding or misuse may cause errors such as input type mismatches (e.g., \textit{"Expected bool, got 1 of type 'int' instead."} in post $\#43604917$\footnote{\url{https://stackoverflow.com/questions/43604917}}) or input data dimension mismatches (e.g., \textit{"Shapes (3, 1) and (1, 3) are incompatible.}'' in post $\#63481755$\footnote{\url{https://stackoverflow.com/questions/63481755}}). 
Listing~\ref{mod_example} shows an example stack trace that belongs to the model-related pattern category. 
We assign the \textit{model training/learning} pattern type to the pattern highlighted in yellow (Line $2$-$6$) in Listing~\ref{mod_example}. We chose this label because the issue that triggers the stack trace is linked with a failure in the model’s learning process caused by the shape mismatch of the logits and labels.
\textbf{Library providers or researchers may help developers avoid or address such issues by providing examples for the APIs or identifying API misuses.}

\nd \textbf{\textit{[DAT] Data-related patterns}}. Pattern types classified in this category focus on the saving, loading, conversion, creation, validation, transformation, and operation of data. 

\begin{lstlisting}[language=Python, label=dat_example, caption={\unexpanded{[DAT]} Data-related patterns: truncated example (ID=$63106341$). The failure is because TensorFlow couldn’t convert a Numpy array into a tensor.}, belowskip=-0.1 \baselineskip]
Traceback (most recent call last):
  File ".../training.py" in _method_wrapper
  File ".../training.py", in fit
  File ".../data_adapter.py",  in train_validation_split
  File ".../nest.py", in map_structure
  File ".../nest.py", in <listcomp>
  File ".../data_adapter.py",  in _split
  File ".../ops.py", in convert_to_tensor_v2
  File ".../ops.py", in convert_to_tensor
  File ".../tensor_conversion_registry.py" in _default_conversion_function
  File ".../constant_op.py", in constant
  File ".../constant_op.py", in _constant_impl
  |\lcolorbox{yellow}{File ".../constant\_op.py" in convert\_to\_eager\_tensor}|
\end{lstlisting}

\nd \textbf{Misunderstanding the data types, data shapes, or the expected input format of an API is the main cause of data-related exceptions.}
Data handling (e.g., data cleaning or data transformation) is essential in developing machine learning applications~\cite{amershi2019software}. Due to the complexity and sheer size of the data, the different input formats of ML algorithms, as well as the variety of data handling APIs (e.g., Numpy or Pandas APIs), developers have difficulties in creating, loading/saving, converting/transforming, or validating their datasets.  
We find that issues related to this category are often related to the misunderstanding of the data types/shapes or the expected input format of an API.
For instance, 
as described in post $\#52832028$\footnote{\url{https://stackoverflow.com/questions/52832028}}, the developer posted a stack trace with an error message \textit{"ValueError: Tensor conversion requested dtype float32 for Tensor with dtype float64: 'Tensor"}. This error occurs when converting a Numpy array into a tensor, as TensorFlow could not convert the \textit{float64} format into the tensor format whereas the \textit{float32} format was expected. 
Listing~\ref{dat_example} shows a stack trace example of the data-related pattern category. 
We identify the pattern (highlighted in yellow (Line $13$)) as a data conversion pattern because the root cause of the failure is related to converting one data type to another (i.e., Failed to convert a NumPy array to a Tensor).
\textbf{Future research or development platforms could help developers identify mismatches in data formats (e.g., by comparing the shape of an input dataset and the expected input data shape of an API).}


\nd\textbf{\textit{[SYN] Python language syntax-related patterns}}. This category reveals failures related to the functionality of the Python programming language. Pattern types in this category include Argument validation, Syntax/attribute extraction, Method wrapper, Python basic syntax, and Object copy.

\begin{lstlisting}[language=Python, label=syn_example, caption={\unexpanded{[SYN]} Syntax-related patterns: truncated example (ID=57717423). The failure is caused By feeding the Dataframe objects instead of Array or Numpy into the one method of the LIME library.}, belowskip=-0.1 \baselineskip]
Traceback (most recent call last):
  File "...\mlp2.py", in <module>
  File "...\mlp2.py", in <listcomp>
  File "...\frame.py", in __getitem__
  |\lcolorbox{yellow}{File "...\textbackslash base.py", in get\_loc}|
  File "...\index.pyx", in pandas._libs.index.IndexEngine.get_loc
  File "...\index.pyx", in pandas._libs.index.IndexEngine.get_loc
\end{lstlisting}

\nd \textbf{Syntax-based exceptions are often caused by the misunderstanding of Python basic syntax such as the type or structure of Python objects.}
Surprisingly, $14.47\%$ of the manually studied patterns are related to Python basic syntax. For example, post $\#56622503$\footnote{\url{https://stackoverflow.com/questions/56622503}} describes an error caused by an incorrect use of the Python built-in module ``random'' when the built-in module was unintentionally overridden by a local file of the same name. This result may be explained by the fact that many ML developers do not have computer science or software engineering training~\cite{braiek2018open}.
In addition, misunderstanding of the type or structure of Python objects is another main cause of this category of stack trace patterns (e.g., as indicated by the Attribute extraction or Object copy stack trace patterns).
Listing~\ref{syn_example} shows an example of the syntax-related stack trace patterns (highlighted in yellow, Line $5$) and assigns the \textit{Syntax/attribute extraction} pattern type to the pattern. The failure was caused by feeding the wrong object type. It means the developer passed \textit{DataFrame} objects instead of \textit{List} or \textit{NumPy} into one method of the LIME library. \textbf{We encourage ML developers to build a solid knowledge of the used programming language to develop their ML applications more effectively and with higher quality.}



\nd \textit{\textbf{[EXT] External dependence related patterns}}. The issues that cause these types of stack traces are related to external environments or outside of the code. This category is composed of: Remote API call, External module extraction, File operation, and Package installation. 

\begin{lstlisting}[language=Bash, label=ext_example, caption={\unexpanded{[EXT]} External dependence-related patterns: truncated example (ID=$55912838$). The failure is related to loading a TensorFlow package and its dependencies on Windows $10$.}, belowskip=-0.1 \baselineskip]
Traceback (most recent call last):
  File "...\pywrap_tensorflow.py" in <module>
  File "...\pywrap_tensorflow_internal.py" in <module>
  File "...\pywrap_tensorflow_internal.py" in swig_import_helper
  |\lcolorbox{yellow}{File "...\textbackslash imp.py" in load\_module}|
  |\lcolorbox{yellow}{File "...\textbackslash imp.py" in load\_dynamic}|
\end{lstlisting}

\nd \textbf{Uncertainties in external dependencies are one of the main causes of the exceptions of ML applications.}
ML applications often rely on external dependencies (e.g., external module/package, remote API call, etc.) to accomplish complex tasks. However, such dependencies may come with uncertainties. For example, an ML application that loads external data may fail when the external data becomes unavailable.  
A manual investigation revealed that many such issues were 
caused by timeout when executing the external dependencies.
For example, in post $\#61149306$, the exception was caused by timeout when loading online package data (i.e., \textit{"socket.timeout: The read operation timed out"}). 
Listing~\ref{ext_example} shows an example of the external dependence related stack trace patterns (Line $5$-$6$, highlighted in yellow). 
We assign the \textit{Package installation} pattern type to the example
as the installation of a specific (incorrect) version of the TensorFlow package caused the error. 
\textbf{Future efforts may help developers mitigate the impact of such problems by pointing them to more robust alternative dependencies.}

\nd 
\textit{\textbf{[MLT] Multi-process related patterns.}} This category reveals failures related to Parallelization and Subprocess invocations. In this category, failures are related to deploying, creating, or executing threads or processes to handle tasks. 

\begin{lstlisting}[language=Python, label=mlt_example, caption={\unexpanded{[MLT]} truncated example (ID=$64116586$). The failure is related to a request for too many parallel jobs.}, belowskip=-0.1 \baselineskip]
Traceback (most recent call last):
  File ".../for_device_data.py" in <module>
  File ".../for_device_data.py" in testPlt
  File "...\dbscan_.py" in fit
  File "...\dbscan_.py" in dbscan
  File "...\base.py" in radius_neighbors
  |\lcolorbox{yellow}{File "...\textbackslash parallel.py" in \_\_call\_\_}|
  |\lcolorbox{yellow}{File "...\textbackslash parallel.py" in dispatch\_one\_batch}|
  |\lcolorbox{yellow}{File "...\textbackslash parallel.py" in \_dispatch}|
  |\lcolorbox{yellow}{File "...\textbackslash \_parallel\_backends.py" in apply\_async}|
  |\lcolorbox{yellow}{File "...\textbackslash \_parallel\_backends.py" in \_\_init\_\_}|
  |\lcolorbox{yellow}{File "...\textbackslash parallel.py" in \_\_call\_\_}|
  File "...\parallel.py" in <listcomp>
  File "...\base.py" in _tree_query_radius_parallel_helper
  File "...\binary_tree.pxi" in sklearn.neighbors.ball_tree.BinaryTree[...]
  File "...\binary_tree.pxi" in sklearn.neighbors.ball_tree.BinaryTree[...]
\end{lstlisting}

\nd \textbf{Multi-process related exceptions in ML applications are often caused by data or resource contentions.} ML processes (e.g., training processes) are usually computationally demanding, thus multi-processing can help in speeding up the processes. An example is the multi-threading\footnote{\url{https://www.tensorflow.org/api_docs/python/tf/config/threading}} feature of TensorFlow. However, multi-processing may lead to data or resource contentions that cause failures.
For example, the post $\#56447556$\footnote{\url{https://stackoverflow.com/questions/56447556}} reports an exception that was triggered by a data contention: multiple processes tried to alter the same data (\textit{"TypeError: can't pickle \_thread.lock objects"}).
This problem could be fixed by disabling multi-processing (e.g., setting the argument \textit{use\_multiprocessing=False}).
Listing~\ref{mlt_example} shows an example of the multi-process related stack trace patterns (Line $7$-$12$, highlighted in yellow). 
We assign the \textit{Parallelization} pattern type to the example 
because the failure was caused due to the creation of a high number of parallel jobs that increased memory consumption until failure. 
\textbf{Future research could help developers identify potential data/resource contentions in their ML applications.}

\nd \textbf{ML exceptions related to external dependencies or manipulations of artifacts (e.g., model or data) are least likely to receive timely community support on SO, indicating their difficulties for developers.}
Table~\ref{tab:table4} illustrates the statistical information of the pattern types with regard to their received community attention and support. For each pattern type, the table shows the number of questions associated with the pattern type, the median number of views of the questions, the median time taken to receive an answer, the percentage of questions that receive an accepted answer, and the median time taken to receive an accepted answer. 
As shown in Table~\ref{tab:table4}, questions with stack traces related to external dependencies (Subprocess invocation\footnote{Although we group the pattern type ``Subprocess invocation'' in the ``Multi-process'' category, it is also related to external dependencies as it involves the invocation of a new and different process, as described in Table~\ref{tab:labels}.}, External module execution, Remote API call, File operation, Package installation) or related to manipulations of artifacts (Data saving, Object copy, Model conversion, Data conversion) are least likely to receive timely community support. Indeed, they are the least likely to receive an accepted answer or take the longest time to receive an accepted answer, or both. Such observations indicate that these exception types are particularly difficult for developers.
 In particular, Model conversion is the most challenging pattern type, with only $21\%$ of the questions receiving an accepted answer. Even when an answer is received, it takes a long time (i.e., a median of $71.46$ hrs) to receive it.  

\input{Table/tbl_rq3_stat.tex}

The difficulties of the exceptions related to external dependencies (e.g., Remote API call) indicate that although the ML developer community on SO may be familiar with ML algorithms, pipelines, and frameworks, they may be less familiar with the hardware/software environment and dependencies that support the development of ML applications. 
For example, in the post $\#42223668$\footnote{\url{https://stackoverflow.com/questions/42223668}}, exception (i.e., \textit{"NotFoundError: Key y\_3 not found in checkpoint"}) was caused by the wrong OS path. Because the user saves the checkpoint inside the working directory but looks for that inside the root path. Indeed, these errors are closer to software engineering rather than machine learning. Experts with mixed backgrounds in machine learning and software engineering could help ML developers answer these questions.

The difficulties of the exceptions related to manipulations of artifacts may be explained by the large size and complex structure of the artifacts. Indeed, machine learning models are becoming increasingly large and complex and are typically treated as black boxes, which makes manipulating such models error-prone (e.g., Model conversion). Similarly, the large size and heterogeneous structure of their data make it challenging to perform manipulations (e.g., Data conversion or Data saving).
For instance, post $\#59078406$\footnote{\url{https://stackoverflow.com/questions/59078406}} wants to convert a pre-trained model into the TensorFlow Lite format. The developer encounters a conversion failure (\textit{"tensorflow.lite.python.convert.ConverterError: TOCO failed."}). This is due to the fact that the \textit{tf.lite.TFLiteConverter} API supports a limited number of ops to be transformed. 
Future work may help ML developers alleviate such challenges by supporting developers in testing/debugging large and complex ML artifacts (e.g., by converting them into smaller or simplified versions for testing/debugging purposes).

\subsection*{\textbf{Discussion}}


\nd\textbf{ML library providers should print actionable error messages when raising exceptions.}
Our manual analysis revealed that many exception error messages lack actionable information for the users. For example, 
in SO post $\#64273829$\footnote{\url{https://stackoverflow.com/questions/64273829}}, a developer posted a stack trace with an error message \textit{"KeyError: <ExtractMethod.NO\_EXTRACT: 1>"} while trying to use TensorFlow Datasets\footnote{\url{https://www.tensorflow.org/datasets}} to load the CelebA dataset. The problem was caused by a Google Drive quota limit (as indicated in the accepted answer).
The developer was unable to figure out the root cause given the stack trace and the error message. The issue could have been resolved more easily (perhaps by the developer itself) with a more informative error message.

\nd\textbf{ML library providers could allow developers to access the templates of the error messages to aid developers in finding relevant forum posts.}
It is known that developers typically search for their issues before posting new questions on SO~\cite{so_search_before}. However, as observed in our manual analysis and prior work~\cite{so_dup}, duplicate posts are common, showing that developers sometimes can not find the answer to their question although a similar question exists. This may be due to the fact that the error messages printed by the libraries often contain dynamic information that is specific to the particular problem faced by a developer. For example, in the error message \textit{"InvalidArgumentError : Incompatible shapes: [32,784] vs. [32,2352]"}, "[32,784]" and "[32,2352]" are dynamic information. If a developer uses the error message directly to search for similar posts, they may miss posts describing the same problem with different dynamic information (e.g., a different data shape). 
To mitigate this problem, library providers could help developers by giving access to error templates (e.g., \textit{"InvalidArgumentError : 
Incompatible shapes: [NUM, NUM] vs. [NUM, NUM]"}). 
Library providers or development platforms could provide a convenient way to copy the templates (e.g., by providing options to copy the original error message or the template). 

\nd\textbf{ML APIs should improve their input argument type and data shape validation.}
In our manual analysis, we found that, for some stack trace patterns, when incorrect data is sent to an API, many internal method calls can be traversed before an error is raised. 
For example, in post $\#39321495$\footnote{\url{https://stackoverflow.com/questions/39321495}}, a developer asks a question about the \textit{"AttributeError: 'list' object has no attribute 'isdigit'"} error. The \textit{pos\_tag()} function expects a \textit{Series} variable as its input but receives a \textit{List}.
Thus, we recommend that ML APIs improve their input argument validation to reduce the spread of erroneous behaviors which can obfuscate 
the original root cause. 


\nd\textbf{ML library providers could incorporate version information in their printed stack traces}. 
In our manual analysis, we found many cases in which there was a mismatch between the code that the user used and the version of libraries they were attempting to use.
For example, in post $\#64771558$\footnote{\url{https://stackoverflow.com/questions/64771558}}, the developer faced an error (\textit{"TypeError: Unexpected keyword argument passed to optimizer: learning\_rate"}) when attempting to convert a Keras model into a Core ML model.
The issue was resolved eventually by ``using the same version of Keras and TensorFlow in the environment creating the model and the environment converting the model''. 
If the version information was indicated in the stack trace (e.g., by printing the version information of the libraries that raise the exceptions), developers might be able to identify the version mismatches with less effort. 
Thus, we recommend ML library providers identify ways to infer or store the signature and versions of APIs that generate stack traces. This information can be used to quickly understand which API belongs to which version of the library and more rapidly determine the root cause of errors.
On the other hand, developers are recommended to carefully verify the versions of the libraries they are using.

\begin{tcolorbox}[enhanced,attach boxed title to top left={yshift=-3mm,yshifttext=-1mm},
  colback=black!5!white,colframe=black!75!black,colbacktitle=gray!80!black,
  title=Summary of RQ3,fonttitle=\bfseries,
  boxed title style={size=small,colframe=black!50!black} ]
We derived five categories of stack trace patterns that are related to models, data, Python syntax, external dependencies, and multi-processing. These exceptions are typically caused by the misunderstanding of ML API usages, data format, or language constructs, or caused by the uncertainties in external dependencies or data/resource contentions in multi-process executions. 
Among them, the exceptions related to external dependencies or manipulations of artifacts (e.g., model or data) are least likely to receive timely community support on SO.
Our observations shed light on future efforts to support ML developers better, for example, to identify misuses of ML APIs, mismatches in data formats, or potential data/resource contentions.
\end{tcolorbox}

%% file: Table/tbl_labels.tex
\begin{table*}[!ht]
\centering
\caption{Classification of 995 stack trace patterns}
\begin{adjustbox}{width=1\textwidth}
\renewcommand{\arraystretch}{1.2} 
\begin{threeparttable}[t]
\begin{tabular}{@{}lllrcr@{}}
\toprule
\multicolumn{1}{l}{\multirow{2}{*}{\textbf{[Category]\tnote{1} \space Pattern Type}}} & \multicolumn{1}{l}{\multirow{2}{*}{\textbf{Description}}} & \multicolumn{1}{c}{\multirow{2}{*}{\textbf{\begin{tabular}[c]{@{}c@{}}Example\\Post\tnote{2}\end{tabular}} }} & \multicolumn{3}{c}{\textbf{Freq.\tnote{3}}} \\ \cmidrule(l){4-6}
\multicolumn{1}{c}{} & \multicolumn{1}{c}{} & \multicolumn{1}{c}{} & \multicolumn{1}{c}{Total} & \multicolumn{1}{c}{(R1, R2)\tnote{4}} & \multicolumn{1}{c}{\textbf{Perc.}} \\\midrule
    \rowcolor{gray!15}
    \textbf{[SYN] Python basic syntax} & This type describes issues with the use of basic syntax in the Python programming language. & 56622503 & 125 & (44, 91) & 14.47\% \\
    \textbf{[MOD] Model training/learning\tnote{5}} & This type of taxonomy covers patterns related to the learning pipeline of an ML model. & 19938587 & 122 & (11, 115) & 14.12\% \\
    \rowcolor{gray!15}
    \textbf{[MLT] Parallelization} & This type indicates patterns about deploying a process across multiple threads or processors.
    & 60905801 & 81 & (23, 64) & 9.38\% \\
    \textbf{[DAT] Data transformation} & This type deals with problems related to the wrong shape, type, or format of the data. & 47499284 & 65 & (8, 59) & 7.52\% \\
    \rowcolor{gray!15}
    \textbf{[MLT] Subprocess invocation} & This type includes patterns for executing a new and different process to handle a new task. & 44523492 & 58 & (23, 42) & 6.71\% \\
    \textbf{[DAT] Data operation} & Patterns come from faulty action manifests during some operations on variables & 44725860 & 52 & (1, 52) & 6.02\% \\
    \rowcolor{gray!15}
    \textbf{[EXT] External module execution} & This type represents patterns related to locating and running External Python modules. & 60705802 & 51 & (25, 33) & 5.90\% \\
    \textbf{[SYN] Method wrapper} & This type includes patterns related to wrapping functions to extend their functionalities. & 60690327 & 49 & (10, 45) & 5.67\% \\
    \rowcolor{gray!15}
    \textbf{[MOD] Model saving/loading} & This type illustrates patterns about loading and saving ML models. & 59757217 & 45 & (2, 44) & 5.21\% \\
    \textbf{[SYN] Syntax/attribute extraction} & This type aggregates patterns relevant to extracting an object's attributes. & 37733245 & 40 & (14, 30) & 4.63\% \\
    \rowcolor{gray!15}
    \textbf{[MOD] Model training/construction} & This type considers patterns affecting building a model during the training process. & 63624390 & 38 & (1, 37) & 4.40\% \\
    \textbf{[DAT] Data saving/pickle} & This type categorizes patterns w.r.t saving a python object as a binary file. & 61916110 & 20 & (14, 10) & 2.31\% \\
    \rowcolor{gray!15}
    \textbf{[DAT] Data validation} & This type includes patterns that are related to the test or validation of a variable. & 63939783 & 19 & (3, 19) & 2.20\% \\
    \textbf{[MOD] Model validation} & It includes patterns according to the testing and validation of a (trained) model. & 43925714 & 18 & (4, 15) & 2.08\% \\
    \rowcolor{gray!15}
    \textbf{[EXT] Remote API call} & This type indicates patterns about establishing and opening HTTP connections. & 59081201 & 14 & (6, 8) & 1.62\% \\
    \textbf{[SYN] Object copy} & This type is for constructing a new compound object. & 44591282 & 14 & (2, 13) & 1.62\% \\
    \rowcolor{gray!15}
    \textbf{[EXT] File operation} & This type is about accessing and manipulating files in an operating system. & 57095189 & 11 & (4, 8) & 1.27\% \\
    \textbf{[EXT] Package installation} & This type is about package installation and uninstallation & 38541233 & 11 & (8, 4) & 1.27\% \\
    \rowcolor{gray!15}
    \textbf{[DAT] Data creation} & It includes the issues arising during the construction of a data object. & 47585817 & 11 & (-, 11) & 1.27\% \\
    \textbf{[MOD] Model conversion} & Problems come up while converting a model to a new one. & 53596521 & 5 & (-, 5) & 0.58\% \\
    \rowcolor{gray!15}
    \textbf{[MOD] Model copy} & This type presents patterns for cloning a model instance. & 54584393 & 4 & (4, -) & 0.46\% \\
    \textbf{[MOD] Hyperparameter tuning} & This type covers patterns about model optimization. & 62205174 & 4 & (1, 3) & 0.46\% \\
    \rowcolor{gray!15}
    \textbf{[SYN] Argument validation} & This type is about checking and meeting the conditions of the input argument of a function. & 63811127 & 3 & (1, 2) & 0.35\% \\
    \textbf{[DAT] Data conversion} & Convert the type or structure of a data to a new one. & 64327397 & 2 & (-, 2) & 0.23\% \\
    \rowcolor{gray!15}
    \textbf{[DAT] Data loading} & This type comprises patterns connected to the decoding and loading of a data object. & 48911850 & 2 & (-, 2) & 0.23\% \\ \bottomrule
\end{tabular}
\begin{tablenotes}
    \item[1] We categorize patterns into 5 different categories, and to save space, we use an abbreviation form for each category. Pattern types that include in one category have the same abbreviation.Categories are: [MOD] Model-related patterns, [DAT] Data-related patterns, [SYN] Python language syntax-related patterns, [EXT] External depending related patterns, and [MLT] Multi-process related patterns. 
    \item[2] For more details about each pattern type, the readers can refer to its web link, “http://stackoverflow.com/questions/”, followed by the post ID. \\(e.g., the example link for the first pattern type is “http://stackoverflow.com/questions/56622503”.)
    \item[3] The frequency column consists of (1) Total: the total frequency of each label for both rounds), (2) (R1, R2): displays the frequency of labels for every round, and (3) Percentage: the share of each label. In the total number, we ignore patterns that are common between rounds. 
    \item[4] R1 and R2 stand for Round 1 and Round 2
    \item[5] Each pattern type may cover distinct aspects. By the slash symbol, we tried to indicate or bold one part of that.

\end{tablenotes}
\end{threeparttable}%
\label{tab:labels}
\end{adjustbox}
\end{table*} 

%% file: Table/tbl_rq3_stat.tex
\begin{table}[!t]
\centering
\small
\renewcommand{\arraystretch}{1.2} 
\caption{Statistical information of pattern types
}
\begin{adjustbox}{width=1\linewidth, center}
\begin{threeparttable}[t]
\begin{tabular}{l||rrrrr}
    \toprule
    \multicolumn{1}{l}{\textbf{[Category]\tnote{1} \space Pattern Type}} & \multicolumn{1}{c}{\textbf{\# Ques.}} & \multicolumn{1}{c}{\textbf{View\tnote{2}}} & \multicolumn{1}{c}{\textbf{\begin{tabular}[c]{@{}c@{}}FAD\tnote{1}\\(Hour)\tnote{2}\end{tabular}}} & \multicolumn{1}{c}{\textbf{\begin{tabular}[c]{@{}c@{}} AAR\tnote{1} \\(\%)\tnote{3}\end{tabular}}} & \multicolumn{1}{c}{\textbf{\begin{tabular}[c]{@{}c@{}}AAD\tnote{1}\\(Hour)\tnote{2}\end{tabular}}} \\
    \hline
    \rowcolor{gray!15}
    \textbf{[SYN] Python basic syntax} & \textbf{1,815} & 423 & 4.72 & 30\% & 4.46 \\
    \textbf{[MOD] Model training/learning} & \textbf{1,733} & 350 & 4.74 & 39\% & 3.51 \\
    \rowcolor{gray!15}
    \textbf{[MLT] Parallelization} & \textbf{1,082} & 480 & 5.40 & 38\% & 4.67 \\
    \textbf{[DAT] Data transformation} & 670 & 252.5 & 3.43 & 38\% & 2.73 \\
    \rowcolor{gray!15}
    \textbf{[MLT] Subprocess invocation} & 542 & 346.5 & \textbf{21.11} & $\bullet$ 28\% & \textbf{23.74} \\
    \textbf{[DAT] Data operation} & \textbf{878} & 356.5 & 9.11 & 36\% & 5.90 \\
    \rowcolor{gray!15}
    \textbf{[EXT] External module execution} & 435 & 269 & 9.01 & $\bullet$ 26\% & 13.05 \\
    \textbf{[SYN] Method wrapper} & \textbf{1,286} & 256 & 4.48 & 32\% & 3.88 \\
    \rowcolor{gray!15}
    \textbf{[MOD] Model saving/loading} & 313 & 359 & 9.62 & 31\% & 6.08 \\
    \textbf{[SYN] Syntax/attribute extraction} & 305 & 206 & 9.58 & 32\% & 7.59 \\
    \rowcolor{gray!15}
    \textbf{[MOD] Model training/construction} & 723 & 508 & 6.46 & 41\% & 4.17 \\
    \textbf{[DAT] Data saving/pickle} & 20 & 266 & 12.33 & 40\% & \textbf{44.28} \\
    \rowcolor{gray!15}
    \textbf{[DAT] Data validation} & 442 & 435 & 2.31 & 39\% & 2.53 \\
    \textbf{[MOD] Model validation} & 116 & \textbf{599} & 2.06 & 40\% & 1.79 \\
    \rowcolor{gray!15}
    \textbf{[EXT] Remote API call} & 34 & 276.5 & 3.36 & $\bullet$ 26\% & 1.61 \\
    \textbf{[SYN] Object copy} & 22 & \textbf{603} & \textbf{79.78} & 32\% & \textbf{127.64} \\
    \rowcolor{gray!15}
    \textbf{[EXT] File operation} & 17 & 394 & 2.48 & $\bullet$ 28\% & 0.52 \\
    \textbf{[EXT] Package installation} & 30 & \textbf{1,312} & 1.60 & 47\% & \textbf{76.22} \\
    \rowcolor{gray!15}
    \textbf{[DAT] Data creation} & 197 & \textbf{531} & 6.91 & 40\% & 4.62 \\
    \textbf{[MOD] Model conversion} & 29 & 214 & \textbf{72.11} & $\bullet$ 21\% & \textbf{71.46} \\
    \rowcolor{gray!15}
    \textbf{[MOD] Hyperparameter tuning} & 19 & 381 & \textbf{20.69} & 42\% & 13.15 \\
    \textbf{[SYN] Argument validation} & 38 & \textbf{1,018} & 2.34 & 37\% & 2.34 \\
    \rowcolor{gray!15}
    \textbf{[DAT] Data conversion} & 46 & 105 & \textbf{23.22} & $\bullet$ 20\% & 11.01 \\
    \textbf{[DAT] Data loading} & 9 & 229 & 6.56 & 44\% & 10.56 \\
    \bottomrule
    \textbf{Overall\tnote{4}} & \underline{251} & \underline{357.75} & \underline{6.51} & \underline{36.5} & \underline{5.285} \\
    \bottomrule
    \end{tabular}
    \begin{tablenotes}
        \item[1] Some column headers are abbreviated due to space constraints. *FAD = First Answer Duration, *AAR = Accepted Answer Rate, *AAD = Accepted Answer Duration
        \item[2] The \textit{View}, \textit{First Answer Duration (FAD)}, and \textit{Accepted Answer Duration (AAD)} columns display the median numbers of each pattern type.  
        \item[3] The \textit{Accepted Answer Ratio (AAR)} column shows the percentage of the total number of questions that received the accepted answer in each pattern type. 
        \item[4] The median of medians
\end{tablenotes}
\end{threeparttable}
\label{tab:table4}
\end{adjustbox}
\end{table}

%% file: Text/threads_to_validity.tex
\section{Threats to validity} 
\label{sec:disc-thre-valid}

\nd We identify the following threats to the validity of our study:
\nd \textbf{External validity threats.} We select SO based on its popularity, importance, and related works.
However, we could increase the generalizability by adding other sources to our research, such as GitHub, for future works. Also, In this study, we focus on posts related to $7$ ML-related libraries. We could further improve generalizability by examining a larger set. In addition, the number of stack trace patterns chosen for Round 1 and Round 2 of our qualitative study may impact the generalizability of our study. To minimize this impact, we select $995$ stack trace patterns using different approaches, such as shared libraries and selective pattern approaches. We assume that our conclusions derived from these $995$ stack trace patterns represent a starting point for the categories and challenges of the stack trace patterns faced by ML developers.

\nd \textbf{Internal validity threats}:  
The stack traces and patterns we use might not wholly indicate developers’ actual challenges. For example, a failure may trigger a stack trace, but the root cause of that failure may be hidden from the stack trace. Also, developers can truncate each stack trace before posting on SO. This concern is more likely to affect patterns that belong to the ML libraries with few stack traces, while patterns with high support can mitigate it.

\nd \textbf{Construct validity threats.} The concordance between SO and SOTorrent is unclear. However, many prior works make use of and draw reasonable conclusions from SOTorrent. In addition, while we use the latest SOTorrent version, it dates from 2020-12-31; at that time, the popularity of some ML libraries was as high as it is currently. Therefore our results may be underestimates. Also, because of the high volume of input data, we cannot check all results individually. Although we perform manual inspections, it is still possible that we may have missed some rare cases that are hard to recognize. Thus, we provide the source code and input data related to this study in our replication package~\cite{replication:Mining:2022}.

%% file: Text/related_works.tex
\section{Related Works}
\label{sec:related-work}



\noindent\textbf{Studies on software stack traces.}
Developers rely heavily on stack traces to find the root causes of issues; hence, much research has been put into the area~\cite{stt8, much1, much2, stack}. Most similar to ours, Medeiros et al.~\cite{stt8} utilize correlation analysis to find and group similar crash reports and identify buggy files in the domain of web-based systems. Also, they assess the performance of the resulting technique in industrial contexts. In this study, we focus on stack traces on SO posts that are related to ML applications. Besides, Sui et al.~\cite{stack} try to build call graphs based on the stack traces collected from the GitHub issue tracker and SO forums. All of the investigations in the paper consider the Java programming language; however, our study is based on the stack traces of ML applications written in Python. 

\noindent\textbf{Studies on ML-related forum posts.}
With a significant increase in the usage of machine learning systems in recent years, a lot of research has been done on the Q\&A forums, such as SO, to understand the state and challenges of ML \cite{alshangiti2019developing, bangash2019developers, islam2019developers, hamidi2021towards, zhang2019empirical}. Alshangiti et al.~\cite{alshangiti2019developing} and Hamidi et al.~\cite{hamidi2021towards} perform empirical studies on SO posts related to ML to find developers’ challenges in developing ML models. They find that most of the questions are related to model deployment phases and that there is a lack of ML experts in the SO community. 
Zhang et al.~\cite{zhang2019empirical} focus on Deep Learning (DL) applications and study the existing challenges in developing DL projects by analyzing the commonly asked questions and their answers on SO. Besides, Bangash et al.~\cite{bangash2019developers} and Islam et al.~\cite{islam2019developers} select posts related to ML to investigate developers’ understanding of ML libraries. The latter focuses on $10$ popular ML libraries such as Tensorflow and Keras and suggests urgently needed research in this area. In another study, Islam et al.~\cite{islam2019comprehensive} analyzed around $2$k posts on SO and $500$ bug fix commits from the GitHub platform with regards to the $5$ DL libraries, including Caffe, Keras, Tensorflow, Theano, and Torch, as well as categorize the characteristics of bugs. 
Similar to the existing studies, our study focuses on SO posts with tags related to $7$ ML libraries. However, different from these studies, we focus on the error patterns of ML applications manifested in stack traces.

\noindent\textbf{Studies on ML library usages.}  
Numerous ML libraries are available for developers to build their ML systems. Understanding the use case of each of these libraries helps developers choose the best fit for their ecosystem. Dilhara et al.~\cite{dilhara2021understanding} conduct an empirical study on the usage of ML libraries by ML developers. They notice (1) a growing tendency to use ML libraries in $2018$ compared to $2013$, and (2) usage of multiple ML libraries in the implementation of ``ML workflows''. Majidi et al.~\cite{AutoML} realize that multiple automated machine learning libraries are infrequently used in scripts and projects. Also, Humbatova et al.~\cite{humbatova2020taxonomy} provide a taxonomy of the existing faults in applications that are based on DL. In addition, Zhang et al.~\cite{Zhang2020} present an empirical study on the DL program failures by focusing on TensorFlow, PyTorch, MXNet, and four related toolkit libraries, including NumPy, DLTK, Detectron, and Fairseq. The input data comes from a DL application in Microsoft, and they manually categorize $400$ failure messages to identify the common root causes of failures. Our work complements the existing studies by studying the SO posts; specifically, seven widespread Python ML libraries, including TensorFlow, Keras, Scikit-learn, PyTorch, NLTK, Hugging Face, and Spark ML, to find patterns in stack traces and their challenges.

%% file: Text/conclusions.tex
\section{Conclusion and future work}
\label{sec:conclusions}
To understand the characteristics of ML stack traces and uncover their patterns and challenges, we perform a large-scale quantitative and qualitative study of the Stack Overflow posts that contain Python stack traces related to seven popular ML libraries. Using this data, we manually identify stack trace pattern types and their categories. Our study reveals that ML stack traces on Stack Overflow (RQ$1$) are common and tend to garner more comments and views than questions without stack traces. However, they are less likely to have accepted answers. We use a common pattern mining algorithm (RQ$2$) to find that recurrent patterns are common in the stack traces of ML questions and that these patterns can be shared across multiple ML libraries. Finally, we manually classify $995$ stack traces into five high-level categories and their challenges and find that misunderstandings in ML API usages, data formats, and language constructs are prevalent causes of errors. Our results can be used to better support ML developers and improve how ML libraries identify errors.

In future work, we plan to create tooling to help users quickly find solutions to their ML-related problems on Stack Overflow, based on our observed stack trace patterns. It is our hope that our results can also help improve how stack traces are used on forums such as Stack Overflow.